\documentclass[10pt,twocolumn,letterpaper]{article}
\usepackage{pdfpages}

\usepackage{cvpr}              %

\usepackage{wasysym}

\usepackage{siunitx}

\definecolor{cvprblue}{rgb}{0.21,0.49,0.74}
\usepackage[pagebackref,breaklinks,colorlinks,citecolor=cvprblue]{hyperref}

\def\confName{CVPR}
\def\confYear{2024}

\title{Polarization Wavefront Lidar: \\Learning Large Scene Reconstruction from Polarized Wavefronts}

\author{First Author\\
Institution1\\
Institution1 address\\
{\tt\small firstauthor@i1.org}
\and
Second Author\\
Institution2\\
First line of institution2 address\\
{\tt\small secondauthor@i2.org}
}

\author{Dominik Scheuble$^{1, 2}$\thanks{These authors contributed equally to this work.} \hspace{0.3em} Chenyang Lei $^{5*}$ \hspace{0.3em} Seung-Hwan Baek$^{4}$ \hspace{0.3em} Mario Bijelic$^{3, 5}$ \hspace{0.3em}  Felix Heide$^{3,5}$ \\
{\small$^1$Mercedes-Benz AG  \hspace{0.3em} $^2$TU Darmstadt  \hspace{0.3em} 
$^3$Torc Robotics   \hspace{0.3em} $^4$POSTECH \hspace{0.3em} $^5$Princeton University}
}

\begin{document}
\maketitle
\begin{abstract}

Lidar has become a cornerstone sensing modality for 3D vision, especially for large outdoor scenarios and autonomous driving. Conventional lidar sensors are capable of providing centimeter-accurate distance information by emitting laser pulses into a scene and measuring the time-of-flight (ToF) of the reflection. However, the polarization of the received light that depends on the surface orientation and material properties is usually not considered. As such, the polarization modality has the potential to improve scene reconstruction beyond distance measurements. In this work, we introduce a novel long-range polarization wavefront lidar sensor (PolLidar) that modulates the polarization of the emitted and received light. Departing from conventional lidar sensors, PolLidar allows access to the raw time-resolved polarimetric wavefronts. We leverage polarimetric wavefronts to estimate normals, distance, and material properties in outdoor scenarios with a novel learned reconstruction method. To train and evaluate the method, we introduce a simulated and real-world long-range dataset with paired raw lidar data, ground truth distance, and normal maps. We find that the proposed method improves normal and distance reconstruction by 53\% mean angular error and 41\% mean absolute error compared to existing shape-from-polarization (SfP) and ToF methods. Code and data are open-sourced \href{https://light.princeton.edu/pollidar/}{here}\footnote{\scriptsize\url{  https://light.princeton.edu/pollidar/}\label{link}}. 
\end{abstract}

\begin{figure*}[!t]
\vspace*{-5mm}
	\centering
	\begin{tabular}{@{}c@{}}
        \includegraphics[width=0.91\linewidth]{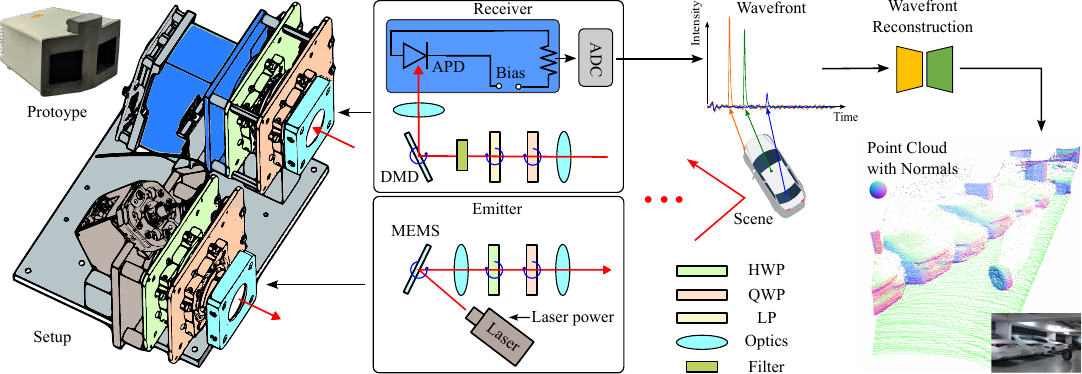}
	\end{tabular}
   \vspace{-2mm}
 	\caption{\textbf{PolLidar sensing and scene reconstruction. }We design our PolLidar sensor with a unique capability: it modulates the polarization of light during both the emission and reception stages. To this end, a HWP and QWP are used to emit light of a certain polarization, whereas a QWP and LP are used to determine the polarization of the received light. To capture the received signal, we employ an ADC at the APD for precise raw wavefront measurement. This is unlike traditional Lidar systems that primarily focus on distance measurements and do not provide both the polarization characteristics and the wavefront of the light. Subsequently, a novel lidar geometry reconstruction approach predicting normals, distance and material properties is introduced in Sec.~\ref{Sec:Reconstruction}.}

	\label{fig:device}
 \vspace{-2mm}
\end{figure*}

\section{Introduction}

Sensing and reconstructing large scenes is crucial for safety-critical applications in autonomous driving~\cite{PV-RCNN,tracking, ZhangLOAM2014}, drones~\cite{kellner2019new, risbol2018lidar}, remote sensing~\cite{dubayah2000lidar, weitkamp2006lidar}, scene understanding~\cite{behley2021towards, geiger20133d, tian2021dynamic} and dataset generation~\cite{carballo2020libre, carlevaris2016university, linnhoff2022simulating} for 3D vision. Scanning lidar sensors have been broadly adopted as a cornerstone sensing modality that provides precise distance information. These sensors operate by measuring the ToF of laser pulses emitted into and returned from the scene. The emitted light is typically polarized and the polarization changes upon reflection depending on surface normals and material properties~\cite{ba2020deep,lei2020polarized}. Off-the-shelf lidar sensors only detect intensity, as such, ignore the additional {polarization} information. 
In this paper, we revisit the abandoned geometric and material information in the polarization state for the reconstruction of large automotive scenes up to 100m range.

Although the benefit of polarization has been investigated extensively in other fields~\cite{lei2020polarized, lyu2019reflection,cui2017polarimetric}, polarization is largely unexplored in the context of lidar sensing in vision and robotics. Specifically, lidar and polarization have been explored in meteorology \cite{sassen2005polarization, sassen1991polarization, schotland1971observations}, biology \cite{huffman2020real} and  
maritime sciences \cite{vasilkov2001airborne} by analyzing the depolarization. Besides, a line of work investigates polarization camera images for shape estimation ~\cite{atkinson2017polarisation_photometric,lei2022shape, riviere2017polarization, berger2017depth, huynh2010shape}, stereo depth estimation~\cite{tian2023dps}, depth completion~\cite{yoshida2018improving}, and dehazing~\cite{Baek:PolarLT:2021, schechner2001instant, fang2014image, treibitz2009active,fuPolarizationSeaFog2023}. These methods have in common that they utilize passive sensors, making them ineffective at night time. Only a few existing works~\cite {Baek:PolarLT:2021,Baek:PolarToF:2022} use active polarimetric ToF systems for scene reconstruction. However, these existing time-resolved polarization methods are designed for indoor scenes with object-level contents, prohibiting the measurement of large outdoor scenes.

\begin{figure}[t]
	\centering
	\begin{tabular}{@{}c@{}}
		\includegraphics[width=1.0\linewidth]{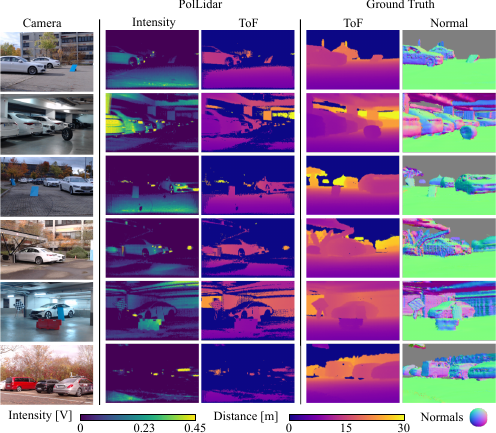}
	\end{tabular}
   \vspace{-2mm}
	\caption{\textbf{PolLidar dataset.} We capture a long-range polarimetric lidar dataset in typical automotive scenes with object distances up to 100m. On the left is a camera reference image, followed by PolLidar intensity for the horizontally polarized state $\theta_1^{\{1,2,3,4\}}=0$ and sensor-derived ToF distances. On the right, ground truth data from accumulated scans from a Velodyne VLS-128 lidar, providing ToF and surface normals for comparison.}
	\label{fig:dataset}
  \vspace{-2mm}
\end{figure}

In this paper, we introduce a novel sensing modality that combines polarization analysis with lidar sensors for scene reconstruction, illustrated in Fig.~\ref{fig:device}. We devise a polarization wavefront lidar sensor (PolLidar) that is capable of operating in outdoor settings. The proposed sensor modulates the polarization of the emitted and received light. In contrast to polarization cameras, the PolLidar is not limited to a discrete number of polarization states but can measure polarization continuously by finely controlling waveplates and linear polarizers basically able to perform full ellipsometry \cite{collett2005field, Goldstein:PolarizedLight}. The sensor reads the raw wavefront signal directly as a voltage from the Avalanche Photodiode (APD). We employ this sensing technique to capture a polarization dataset consisting of long-range automotive scenes to assess the benefit of polarization. Along with the raw wavefronts, we provide pairwise ground truth distance and normal information from a Velodyne VLS-128 reference sensor, see~Fig.~\ref{fig:dataset}. 

To recover scene properties from polarization wavefront measurements, we combine the proposed sensor with a novel reconstruction approach that operates on the raw polarimetric wavefronts. The proposed reconstruction method uses the polarized wavefronts to estimate surface normals and accurate distance. The estimated normals can then be utilized for predicting material properties, including index of refraction, diffuse and specular albedo, and surface roughness. For training, we extend the CARLA simulator~\cite{goudreault2023lidarhyperoptim} with a realistic polarization model of light to generate a synthetic long-range polarization dataset. 

We assess the method with experiments on both synthetic data and real-world data. We find that the proposed method improves distance estimation by 41\% mean absolute error compared to conventional ToF methods and 53\% mean angular error for normal estimation compared to SfP and point cloud baselines on automotive scenes.

\vspace{5pt}\noindent
Specifically, we make the following contributions
\begin{itemize}
    \item We devise a polarization wavefront lidar sensing approach that measures time-resolved polarization properties to recover precise distance and normals for long-range scenarios as found in automotive scenes.
    \item We propose a neural reconstruction approach for distance and normals operating directly on raw wavefronts instead of post-processed ToF peaks.
    \item We introduce the first automotive polarization lidar dataset, consisting of real-world data and simulation data. We validate our model with the proposed dataset for long-range distance estimation and dense normal reconstruction. Compared to baseline methods, our model improves distance and normal reconstruction by {41}\% mean absolute error and {53}\% mean angular error, respectively.
\end{itemize}

\section{Related Work}
\noindent\textbf{Polarization Lidars.} Polarization lidar sensors have been explored in diverse fields. Early studies, such as Schotland's~\cite{schotland1971observations}, leveraged these polarimetric measurements for cloud property analysis, while approaches as ~\cite{schotland1971observations} study the bioaerosols in the atmosphere~\cite{huffman2020real} and in \cite{vasilkov2001airborne}  the scattering coefficient of oceans are measured using polarization lidar~\cite{vasilkov2001airborne}. Recently, Baek et al.~\cite{Baek:PolarToF:2022,Baek:PolarLT:2021} combine a prototypical polarization lidar with a temporal-polarimetric BRDF model to achieve accurate scene reconstruction. Jeon et al.~\cite{jeon2023polarimetric} propose a polarimetric indirect ToF imaging method that utilizes polarization to improve depth estimations through scattering media. 
However, the imaging technique, i.e., the design of the optical path in ~\cite{Baek:PolarToF:2022,Baek:PolarLT:2021}, and the indirect ToF measurement principle in~\cite{jeon2023polarimetric}, fundamentally limit these devices to indoor usage. In contrast, the proposed method is the first designed for scene reconstruction in large outdoor scenes up to 100m.

\noindent\textbf{Scene Reconstruction with Passive Polarization Sensors.} 
Exploiting the relationship between the polarization of reflected light and the surface normals, shape from polarization (SfP) methods have achieved scene reconstruction from polarization images captured by linear-polarization cameras~\cite{rahmann2001reconstruction,atkinson2006recovery,miyazaki2003polarization}.  Early SfP methods focus on estimating the surface normal of objects under assumptions of either pure specular reflection~\cite{rahmann2001reconstruction} or pure diffuse reflection~\cite{atkinson2017polarisation_photometric,miyazaki2003polarization}. These methods usually assume an unpolarized light source and suffer from polarization ambiguity issues. Recent works~\cite{ba2020deep,kondo2020accurate,lei2022shape,ikemura2024robust} leverage deep learning to solve the ambiguity problem. By training on real-world datasets, the network can better distinguish the ambiguity and mitigate the need for inputting unknown material properties such as refractive index. 
Baek et al.~\cite{baek2018simultaneous} perform joint optimization of appearance, normals, and refractive index. Deschaintre et al.~\cite{deschaintre2021deep} propose a learning-based inverse learning framework with the front-flash illumination. Dave et al.~\cite{dave2022pandora} combine polarization with implicit neural representations to collectively reconstruct the geometry and appearance from multiple images. 
In general, these reconstruction methods focus on scenes with few objects that are placed to exhibit strong polarization cues with a high degree of polarization (DoP).
In outdoor scenes, however, the DoP varies significantly limiting the quality of the reconstruction to high DoP regions. The proposed method allows to exploit the exploitation of polarization cues in both high and low DoP regions.

In \cite{kadambi2015polarized,yoshida2018improving}, passive polarization sensors are combined with other imaging modalities. Kadambi et al.~\cite{kadambi2015polarized} utilize normals from polarization to enhance the details of depth from a Microsoft Kinect sensor. Yoshida et al.~\cite{yoshida2018improving} use polarization to fill in missing regions in the depth maps. Furthermore, polarization cues are leveraged to augment low-quality depth maps from two-view stereo~\cite{zhu2019depth,fukao2021polarimetric}, reciprocal image pairs~\cite{ding2021polarimetric}, multi-view stereo~\cite{cui2017polarimetric, Miyazaki2016SurfaceNE}, or lidars~\cite{shakeri2021polarimetric}. Recently, Huang et al.~\cite{huang2023learning} and Tian et al.~\cite{tian2023dps} propose stereo polarimetric methods, which utilize two polarization images to solve the ambiguity in SfP. However, as passive sensors are dependent on ambient light, these methods struggle in low-light conditions. The proposed active sensing method allows for accurate reconstructions independently of ambient illumination.

\section{Polarimetric Wavefront Lidar}

\begin{figure*}[!t]
	\centering
	\includegraphics[width=0.95\linewidth]{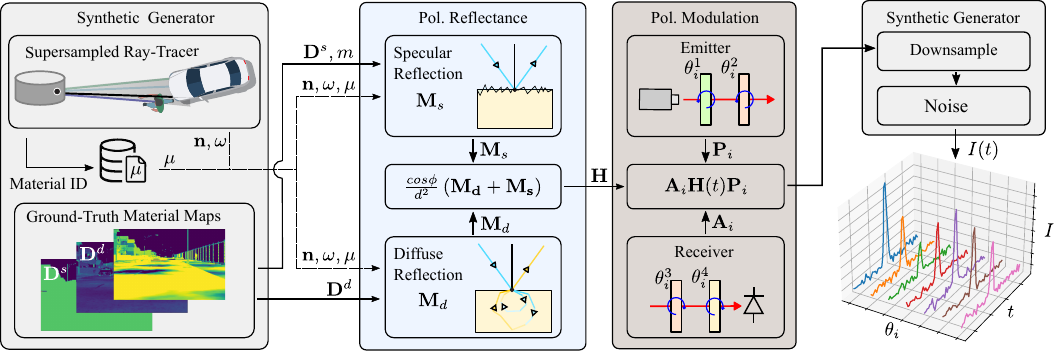}
  \vspace{-2mm}
	\caption{\textbf{PolLidar forward model and simulator.} Temporal polarimetric reflectance of the scene can be modeled as the sum of specular $\mathbf{M}_s$ and diffuse $\mathbf{M}_d$ reflection. Receiver and emitter of the PolLidar can be described with the Mueller matrices $\mathbf{P}_i$ and $\mathbf{A}_i$ that are functions of the rotation angles $\theta_i^{\{1,2,3,4\}}$ of HWP, QWPs and LP, respectively. We employ the resulting PolLidar forward model in a simulator based on CARLA that generates synthetic polarimetric raw wavefronts. To this end, we extract material properties and normals from CARLA and feed them into the forward model. The resulting temporal wavefronts are subsequently downsampled in spatial dimension to model beam divergence and noise is added to simulate APD and ADC.} 
 	\label{fig:carla}
    \vspace{-3mm}
\end{figure*}

In environmental science,  polarimetric lidars are employed for gathering polarization data over extensive ranges, often spanning several kilometers but with a trade-off in spatial resolution. Contrarily, polarimetric lidars for scene reconstruction usually support high spatial resolution, yet their range is limited to a few meters. The proposed PolLidar sensor in Fig.~\ref{fig:device} uniquely bridges these application domains. It is designed to allow for a balanced performance optimal for both long-range capabilities up to 223 m and high spatial resolution of 150 rows and 236 columns over a 23.95\textdegree\ and 31.53\textdegree\ vertical and horizontal field-of-view, making it particularly suitable for autonomous driving applications. %

Our sensor differs from the ToF systems described in \cite{Baek:PolarToF:2022, Baek:PolarLT:2021}. Specifically, we propose separate modules for emission and reception instead of a shared optical setup. This separation allows for a larger optical aperture in each module, enhancing optical sensitivity and extending the operational range in outdoor scenarios.
Instead of the galvo-mirror used in \cite{Baek:PolarLT:2021}, a MEMS micro-mirror is used in the emitter for scene scanning. 
The receiver employs a digital micro-mirror device (DMD) following \cite{royo2012lightbeam} to selectively deflect the returning light towards the photodiode.
Using the DMD allows for apertures ($\diameter$0.55") comparable to galvo-mirrors effectively reducing light loss.

To make outdoor applications possible, we operate at a wavelength of 1064 nm and added a narrow bandpass filter, leaving the visible band targeted by existing work. These modifications are essential to suppress ambient light and render the emitted light invisible to the human eye, aligning with automotive illumination standards. The maximum power output adheres to Class-1 eye safety regulations.
The laser power remains adjustable according to scenario requirements, offering a balance between achieving maximum range and minimizing saturation which offers a level of control typically not available in off-the-shelf lidars.

On the emission side, the horizontally polarized laser light undergoes modulation by passing through both a half (HWP) and quarter-wave plate (QWP). The receiving module is designed to capture changes in polarization, facilitated by a sequence of a QWP, a linear polarizer (LP), and a bandpass filter, as illustrated in Fig.~\ref{fig:device}. The rotation of each polarization element is finely adjustable in increments of 0.01 degrees.
We use a back-side illuminated Avalanche Photodiode (APD) with an adjustable bias for sensitivity adjustments and read the raw signal with an attached PCIe-5764 FlexRIO-Digitizer analog-to-digital converter (ADC), sampling at 1 Gs/s. This allows us to measure raw wavefronts with a length of 1488 bins of 1~ns width, i.e., 15 cm per bin, and a range of 223 meters. 

Our prototype design is optimized for a highly configurable selection of polarization states by finely controlling the polarization elements at the expense of measurement time.
The acquisition of a frame, as described in Sec.~\ref{subsec:proprecess}, results in a capture time of 5~min.
We refer to the Supplementary Material for an analysis on how future setups can achieve real-time capability.
Although adding polarization requires additional complexity, we argue that the potential benefits extend beyond the scope of this paper, aiding reconstruction in scenarios with multi-path reflections or scattering media \cite{Baek:PolarLT:2021}.

\subsection{Polarimetric Lidar Forward Model}
We model polarization with the Stokes-Mueller formalism, with light and reflectance described by a Stokes vector $\mathbf{s} \in \mathbb{R}^{4\times1}$ and a Mueller matrix $\mathbf{M} \in \mathbb{R}^{4\times4}$.~\cite{collett2005field,baek2020image}.
Recently, Baek et al.~\cite{Baek:PolarToF:2022} introduced a temporal-polarimetric reflectance model $\mathbf{M}(\tau, \boldsymbol{\omega}_i, \boldsymbol{\omega}_o)$ describing how light polarization and intensity change when impinging on a surface with given incident and outgoing direction of light ($\boldsymbol{\omega}_i$ and $\boldsymbol{\omega}_o$), and with temporal delay ($\tau$) of diffuse reflection. 
As shown in Fig.~\ref{fig:carla}, the reflectance $\mathbf{M}$ can be modeled as a sum of specular and diffuse reflection ($\mathbf{M}_s$ and $\mathbf{M}_{d}$)
\begin{align}\label{eq:tpBRDF}
\mathbf{M}\left(\tau, \boldsymbol{\omega}_i, \boldsymbol{\omega}_o\right) &= \mathbf{M}_s(\tau, \boldsymbol{\omega}_i, \boldsymbol{\omega}_o) + \mathbf{M}_{d}(\tau, \boldsymbol{\omega}_i, \boldsymbol{\omega}_o) \\
\mathbf{M}_s\left(\tau, \boldsymbol{\omega}_i, \boldsymbol{\omega}_o\right) &= \frac{D(\theta_h; m)G(\theta_i, \theta_o; m)}{4\cos\theta_i \cos\theta_o} \mathbf{D}^s(\tau) \mathbf{F}_R\\
\mathbf{M}_{d}\left(\tau,\boldsymbol{\omega}_i, \boldsymbol{\omega}_o\right) &= \mathbf{C}_{\mathrm{n}\rightarrow\mathrm{o}}\mathbf{F}_T^o\mathbf{D}^{d}(\tau)\mathbf{F}_T^i\mathbf{C}_{\mathrm{i}\rightarrow\mathrm{n}},
\end{align}
where $\theta_h = \cos^{-1}(\mathbf{h} \cdot \mathbf{n})$, $\theta_i = \cos^{-1}(\mathbf{n} \cdot \boldsymbol{\omega}_i)$, $\theta_o = \cos^{-1}(\mathbf{n} \cdot \boldsymbol{\omega}_o)$ and $\mathbf{n}$ is the surface normal.
$D$ and $G$ are functions to describe the surface, where $m$ is the roughness. $\mathbf{C}_{\mathrm{i}\rightarrow\mathrm{n}}$ and $\mathbf{C}_{\mathrm{n}\rightarrow\mathrm{o}}$ are the coordinate-conversion Mueller matrices~\cite{collett2005field}, and $\mathbf{F}_T^i$, $\mathbf{F}_T^o$ are the Fresnel transmission Mueller matrices for incident and outgoing light, depending on refractive index $\eta$. $\mathbf{D}^{s}$ and $\mathbf{D}^{d}$ are the depolarization Mueller matrices for specular and diffuse reflections ~\cite{Baek:PolarToF:2022}.

Given its long-range working distance, we can assume that the incident and outgoing direction of light in our sensor are identical, and we approximate the reflectance model \eqref{eq:tpBRDF} with a single viewing direction $\boldsymbol{\omega} = \boldsymbol{\omega}_i = \boldsymbol{\omega}_o$.
After scaling $\mathbf{M}$ by the cosine shading term $\cos  \phi$  and attenuation such that $\mathbf{H} \left(\tau, \boldsymbol{\omega},\boldsymbol{\omega}  \right) = \left( \cos  \phi \mathbin{/} d^2 \right) \mathbf{M} \left(\tau, \boldsymbol{\omega}, \boldsymbol{\omega} \right)$, the lidar forward model can be written as
\begin{align}
\label{eq:image_formation_tof}
  I(t, \boldsymbol{\omega}) &= \left[\int_{0}^{t'}\mathbf{H}(\tau, \boldsymbol{\omega}, \boldsymbol{\omega}) \mathbf{s}_\textrm{laser}(\boldsymbol{\omega}, t'-\tau)\mathrm{d}\tau\right]_0,
\end{align}
where $t' = t-2d/c$, $d$ is the distance between laser and scene, and $c$ is the speed of light.
$\mathbf{s}_\textrm{laser}$ denotes the Stokes vector of the emitted laser light.
The operator $\left[...\right]_0$ denotes taking the first element of the resulting vector.

\begin{figure*}[!t]
	\centering
	\includegraphics[width=0.95\linewidth]{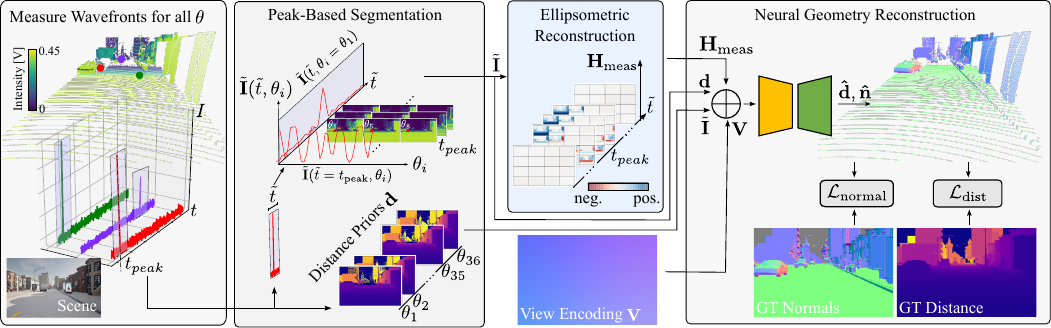}
  \vspace{-2mm}
	\caption{\textbf{Neural polarization wavefront lidar reconstruction.} We capture raw polarization wavefronts of the scene $\mathbf{I}$. We apply a peak-based segmentation technique to obtain a sliced polarization wavefront $\Tilde{\mathbf{I}}$ and distance priors $\mathbf{d}$. Via ellipsometric reconstruction, we estimate a sliced Mueller matrix $\mathbf{H}_{\mathrm{meas}}$. Finally, we concatenate all the polarization priors with viewing direction $\mathbf{V}$ as the input to a neural network predicting distance and normals for the scene. We supervised the network with a normal loss $\mathcal{L}_{\mathrm{normal}}$ and a distance loss $\mathcal{L}_{\mathrm{dist}}$.
}
	\label{fig:Architecture}
 \vspace{-3mm}
\end{figure*}

We use rotating ellipsometry to infer all elements of the Stokes vectors~\cite{collett2005field}. 
As illustrated by Fig.~\ref{fig:carla}, a HWP and a QWP are rotated to modulate the polarization of the emitted light.
Analogous on the receiving side, a QWP and a LP are used to measure light with a specific polarization incident on the APD.
Hence, the image formation of the PolLidar can be modelled as
\begin{equation}\label{eq:ellipsometry}
  I_i(t, \boldsymbol{\omega}) = \left[\mathbf{A}_i \mathbf{H}(t', \boldsymbol{\omega}) \mathbf{P}_i\mathbf{s}_\textrm{laser}(\boldsymbol{\omega},0)\right]_0,
\end{equation}
where $\mathbf{A}_i$ and $\mathbf{P}_i$ are the $i$-th Mueller matrices of the analyzing optics and the polarizing optics defined as
  $\mathbf{A}_i = \mathbf{L}(\theta_i^4)\mathbf{Q}(\theta_i^3)$ and $\mathbf{P}_i = \mathbf{Q}(\theta_i^2)\mathbf{W}(\theta_i^1),$
with $\theta_i^{\{1,2,3,4\}}$ as the rotation angles of the emitter HWP and QWP and the receiver QWP and LP, respectively. $\mathbf{W}$, $\mathbf{Q}$, and $\mathbf{L}$ are the Mueller matrices of the HWP, QWP, and LP~\cite{collett2005field}.
The integral is omitted as a result of using pulsed laser illumination.

\subsection{Polarimetric Lidar Simulator}

In order to use the PolLidar in a learning-based framework, a sufficient amount of training data is required.
However, the finely controllable polarization elements come at the cost of longer measurement times as the motors move relatively slow. 
To acquire a large polarization wavefront dataset, we integrate the lidar forward model from Eq.~\eqref{eq:ellipsometry} into the CARLA simulator \cite{dosovitskiy2017carla} to generate vast amounts of synthetic training data.
Specifically, we extend the full wavefront lidar model for CARLA as introduced by \cite{goudreault2023lidarhyperoptim}.
As presented in Fig.~\ref{fig:carla}, we extract the material properties $m$, $\mathbf{D}^s$ and $\mathbf{D}^d$  using custom material cameras.
However, materials in CARLA do not have refractive indices $\mu$ assigned by default.
We circumvent this problem by extending the ray-tracer to return the material ID of each hit point.
Based on the material ID, we look-up the corresponding refractive index $\mu$ in a database \cite{iordatabase}.
Additionally, we extend the ray-tracer to return normals $\mathbf{n}$ for each hit point.

With material properties and normals in hand, we simulate the scene using the polarimetric lidar forward model.
To model the beam divergence of the laser beam, we downsample neighboring rays to eventually render the temporally resolved polarimetric raw wavefronts.
Next, we model shot and read-out noise by applying Poisson and Gaussian noise to the wavefronts, respectively.
We tune the noise characteristics such that they closely resemble the real device.
Additional details are provided in the Supplementary Material.

\begin{figure*}[!t]
	\centering
	\begin{tabular}{@{}c@{}}
		\includegraphics[width=1.0\linewidth]{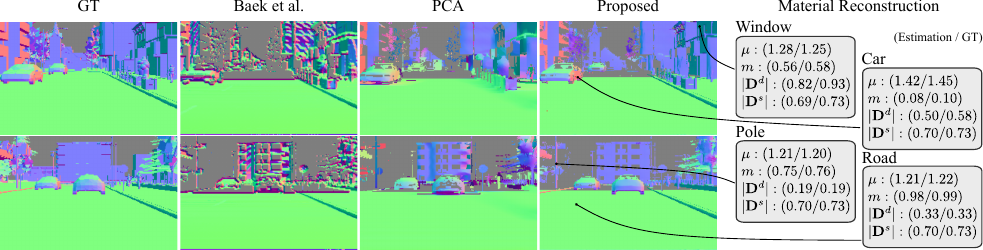}
	\end{tabular}
\vspace{-2mm}
\caption{\textbf{Qualitative evaluation on synthetic data.} Baek et al.~\cite{Baek:PolarToF:2022} is unable to reconstruct normals in areas with low DoP, e.g., walls of buildings facing the sensor. PCA~\cite{Zhou2018open3d} applied in this setting are strongly dependent on point cloud density. Thus, distant poles and cars in the second row cannot be reconstructed accurately. The proposed approach leverages polarization cues to reconstruct normals in sparse regions and is robust against low DoP areas. We estimate accurate material properties for different surfaces and objects (right).}
\label{fig:baselines}
\vspace{-1em}
\end{figure*}

\section{Neural Polarization Lidar Reconstruction}\label{Sec:Reconstruction}
To leverage polarized raw wavefront data, we devise a learning-based approach for reconstructing normals and distance as presented by Fig.~\ref{fig:Architecture}. First, we preprocess the wavefronts as described in Sec.~\ref{subsec:proprecess}. 
Next, we train a neural network to predict normals and distance from polarized wavefronts as discussed in Sec.~\ref{subsec:NeuralReconstruction}.

\subsection{Preprocessing Wavefronts} 
\label{subsec:proprecess}

When capturing a frame, we perform rotating ellipsometry by collecting raw wavefronts for 36 different rotation angles $\theta_i$ subsequently denoted as $\mathbf{I} = \{I_i\}_{i=1}^{36}$, where  $ I_i \in \mathbb{R}^{H \times W \times T}$ with $H{=}150$, $W{=}236$ and $T{=}1488$. The temporal resolution $T$ and the repeated measurement for each angle $\theta_i$ results in 53,568 samples for each ray in $\mathbf{I}$. To tackle this large dimensional space, we first perform peak-based segmentation to obtain sliced wavefronts as shown in Fig.~\ref{fig:Architecture}.
Specifically, to reduce the temporal dimension, we first locate the peak within the wavefront. Then, we segment a window of size 51 centered around the peak, resulting in a sliced wavefront $\mathbf{\Tilde{I}} = \{\tilde{I}_i\}_{i=1}^{36}$, where $\tilde{I}_i \in \mathbb{R}^{H \times W \times 51}$. 
We preserve the temporal index of the peak $t_{\mathrm{peak}}$ as it contains the distance information ${\mathbf{d}}\in \mathbb{R}^{36 \times H \times W \times 1}$. 

As the raw wavefront $\mathbf{\Tilde{I}}$ implicitly encodes the polarization optics from emitter and receiver, we apply ellipsometric reconstruction to recover the time-dependent Mueller matrix $\mathbf{H}$. To this end, we use the temporal measurements $I_i$ collected at various rotation angles of the polarizing optics to invert the image formation model presented in Eq.~\eqref{eq:ellipsometry}. Following the approach of Baek et al.~\cite{baek2020image}, we recover the Mueller matrix $\mathbf{H}_\mathrm{meas} \in \mathbb{R}^{H \times W \times 51 \times 16}$ by solving a least-squares optimization problem as follows
\begin{equation}\label{eq:M_meas}
\mathop {{\rm{minimize}}}\limits_{\mathbf{H}_\mathrm{meas}} \sum_{i=1}^N
\left(I_i - \left[ \mathbf{A}_i \mathbf{H}_\mathrm{meas} \mathbf{P}_i \mathbf{s}_\textrm{laser} \right]_{0}\right)^2.
\end{equation}

\subsection{Neural Geometry Reconstruction} 
\label{subsec:NeuralReconstruction}
Subsequent to the pre-processing, we reconstruct the geometry of the scene by inputting the signals to a neural reconstruction network. 
To this end, the temporal dimension is flattened and the all inputs are concatenated as input $\mathbf{x}$
\begin{equation}
    \mathbf{x} = \Tilde{\mathbf{I}} \oplus \mathbf{d} \oplus \mathbf{H}_\mathrm{meas} \oplus \mathbf{V},
    \label{eq:ddpm_concat}
\end{equation}
where $\oplus$ denotes concatenation along the feature dimension and $\mathbf{V}$ is the viewing direction. 
We then predict normals $\hat{\mathbf{n}}$ and distance $\mathbf{\hat d}$  with a neural network. The network is a variation of a TransUnet that combines the U-Net and transformer architecture components. Specifically, we use 3 encoder layers to encode the features. At the bottleneck, we use 8 transformer layers. At last, we use 3 decoder layers with skip-connection to predict normals and distance.

To train the  network, we supervise normals and distance predictions with a cosine similarity loss for the surface normals and a mean absolute loss for distance
\begin{align}
        \mathcal{L}_\mathrm{normal} &= |1 - (\mathbf{c} \odot \mathbf{n}_\mathrm{gt} ) \cdot (\mathbf{c} \odot \hat{\mathbf{n}}))|_1, \\
        \mathcal{L}_\mathrm{dist} &= |\mathbf{c} \odot \mathbf{d}_\mathrm{gt} - \mathbf{c} \odot \hat{\mathbf{d}})|_1,
\end{align}
where $\mathbf{c}$ is the confidence mask for the normals where ground truth normals are not available.

We implement the proposed method in PyTorch. We train the model for 200 epochs on a Nvidia A100 GPU. We use the Adam optimizer with a learning rate of 1e-4 and we set the batch size to 1. We crop images to 128×128 patches in each iteration for augmentation. We apply different laser powers and biases during training to increase robustness against saturation and low-intensity readings. More details are presented in the Supplementary Documentation.

\begin{table}[!t]
\small
\centering
\renewcommand{\arraystretch}{1.2}
\resizebox{1.0\linewidth}{!}{
\begin{tabular}{l@{\hspace{5mm}}c@{\hspace{3mm}}c@{\hspace{3mm}}c@{\hspace{5mm}}c@{\hspace{3mm}}c@{\hspace{3mm}}c}
\toprule[1pt]
\small{Method}&\multicolumn{3}{c}{Angular Error [\textdegree] $\downarrow$ } & \multicolumn{3}{c}{Accuracy [\%] $\uparrow$}  \\ 
 & \small{Mean} & \small{Median}  & \small{RMSE}& \small{$3.0^{\circ}$} & \small{$5.0^{\circ}$} & \small{$10.0^{\circ}$} \\
 
\midrule[0.6pt]

SfP-DoP~\cite{atkinson2006recovery} &49.82 &	35.00	&65.39&	4.29&	7.60&	12.63  \\
Baek et al.~\cite{Baek:PolarToF:2022} &31.03&8.32&	53.21&	27.12	&44.44&	61.03\\
PCA~\cite{Zhou2018open3d} & 18.64&	8.02&	33.60	&55.89&	60.64&	66.84\\
Proposed       & \textbf{8.71} &	\textbf{4.31}	 &\textbf{17.65} &	\textbf{65.49} &	\textbf{70.19}	 &\textbf{78.15}       \\ 
\bottomrule[1pt]
\end{tabular}
}
\vspace{-2mm}
\caption{ \textbf{Quantitative evaluation for normals on synthetic data.} The SfP baseline~\cite{atkinson2006recovery} is unable to reconstruct normals in real-world as the underlying assumptions do not translate to real-world scenarios. Baek et al.~\cite{Baek:PolarToF:2022} is designed for object-level ToF imaging and fails in low DoP regions. PCA~\cite{Zhou2018open3d} achieves improved results but with quality depending on point cloud density. The proposed method leverages both the neighborhood of points and the polarization cues; thus outperforming all the baselines.}
\label{table:baseline_cmp_syn}
\vspace{-1em}
\end{table}

\section{Assessment}

To assess the effectiveness of the proposed reconstruction method, we first validate the method on synthetic data with perfect ground truth. Next, we discuss material estimation before validating the method with the experimental device. Finally, we ablate the different inputs to show the benefit of polarized raw wavefronts.

\begin{figure*}
	\centering
	\begin{tabular}{@{}c@{}}
		\includegraphics[width=1.0\linewidth]{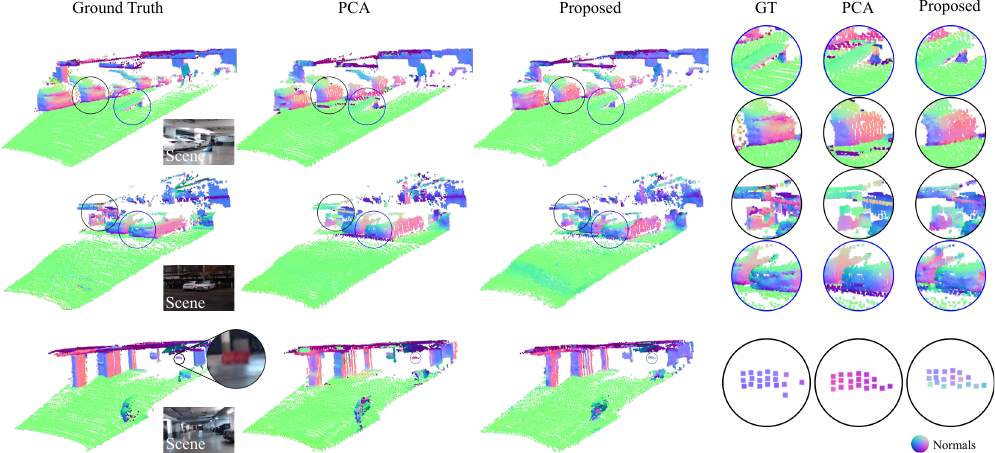}
	\end{tabular}
 \vspace{-2mm}
\caption{\textbf{Qualitative evaluation on experimental data.} PCA~\cite{Zhou2018open3d} applied on measured captures from our PolLidar results in erroneous predictions of surface normals, especially prominent for the fine structures visible in the zoom-ins of the first two rows, see. e.g., the transition of ground and metal ramp in the first row and the metal roof support in the second row. In contrast, the proposed method is able to resolve these fine details. In the last row, we show a lost cargo scenario with an upright object blocking the road in 50m distance. Our method correctly classifies the object as facing toward the vehicle, whereas PCA predicts a flat surface with downwards oriented normals.}
\label{fig:baselines_real}
\vspace{-1.5em}
\end{figure*}

\subsection{Synthetic Evaluation}

We first validate the proposed neural geometry reconstruction method on synthetic data with perfect ground truth. We compare against three SfP baseline methods to evaluate the quality of the reconstructed normals. Specifically, we evaluate against Baek et al.~\cite{Baek:PolarToF:2022} as a baseline designed for object-level scene reconstruction for a polarimetric ToF prototype. This approach fits the recovered Mueller matrix $\mathbf{H}_\text{meas}$ to the polarimetric lidar forward model by jointly estimating material properties and normals. Next, we compare against the classical SfP approach from \cite{atkinson2006recovery}, which recovers surface normals from the DoP by assuming a scene-wide constant refractive index and diffusive reflection.
As reported in Tab.~\ref{table:baseline_cmp_syn}, classical SfP approaches do not generalize well to outside scenes.
This can be attributed to real-world geometry exhibiting regions of high but also very low DoP.
Low DoP regions occur when the surface normal and the viewing direction of the lidar align, see Supplementary Dcoumentation. Highlighted by the qualitative findings in Fig.~\ref{fig:baselines}, the method from Baek~et~al.~\cite{Baek:PolarToF:2022} is unable to reconstruct normals in low DoP regions, e.g., buildings of walls that face the sensor, whereas for high DoP regions, as e.g., the side of a vehicle, satisfying performance is achieved.

Moreover, we compare against conventional lidar by averaging wavefronts from all polarization states and applying peak-finding.
Treating our PolLidar as conventional makes for an adequate comparison as the number of scanned points is equal for both conventional and PolLidar.
Subsequent for normal reconstruction, we apply PCA~\cite{Zhou2018open3d} as a point-cloud based method that considers a neighborhood of points.
This method performs well in areas with flat geometry and high point density but degrades significantly at long ranges with sparse distance, e.g., cars in far distances in the second row of Fig.~\ref{fig:baselines} and geometry transition regions, e.g., the area between road and car. 
PCA also struggles with thin structures like the pole in the second row of Fig.~\ref{fig:baselines}. The proposed method leverages the additional polarization cues to resolve normals in regions with sparse points. We also achieve satisfying reconstruction results for regions with weak polarization information by taking a local neighborhood and cues from a normal-dependent widened pulse into account. 
As a result, the proposed approach outperforms PCA~\cite{Zhou2018open3d} by 53\% on the mean angular error as shown in Tab.~\ref{table:baseline_cmp_syn}.

For evaluating distance estimation, we compare against the conventional argmax-peak-finding typically performed directly on the device by low-level electronics \cite{behroozpour2017lidar}.
This approach is limited by the temporal resolution of the sensor and we find a mean absolute distance error of 32cm.
The proposed method leverages the raw wavefront data and the relationship between distance and normals to generate high-quality distance.
We find that the proposed method yields a mean absolute distance error of 19cm outperforming the conventional approach by {41}\% mean absolute error. 
Additional metrics are provided in the Supplementary Material.

\subsection{Material Property Estimation}

With estimated surface normals in hand, we reconstruct the material properties, namely index of refraction $\mu$, roughness $m$ and the depolarization matrices $\mathbf{D}^s$ and $\mathbf{D}^d$, of the polarimetric lidar forward model.
To this end, we follow Baek~et~al. \cite{Baek:PolarToF:2022} and estimate material properties by rendering the Mueller matrix $\mathbf{H}_\text{render} = \mathbf{H}^s_{\text{render}} + \mathbf{H}^d_{\text{render}}$ that best explains the reconstructed Mueller matrix $\mathbf{H}_\text{meas}$.
In the large scenes we tackle, we find that the DoP is mostly governed by diffuse reflection. 
We leverage this heuristic to disentangle the specular and diffusive Mueller matrices.
To this end, we solve the following minimization problem
\begin{equation}
    \begin{aligned}
    \mathop {{\rm{minimize}}}\limits_{\mu, m, |\mathbf{D}^s|, |\mathbf{D}^d|}~
    & \lambda^d  \left|  \mathbf{c}_\text{dop} \odot (\mathbf{H}_\text{meas} - \mathbf{H}^d_\text{render}) \right|_1 + \\
    &\lambda^s \left|(\mathbf{H}_\text{meas} - \mathbf{H}^d_\text{render}) -  \mathbf{H}^s_\text{render}\right|_1,
    \end{aligned}
\end{equation}
where $\lambda^d$, $\lambda^s$ are scalar weights and $\mathbf{c}_\text{dop}$ a mask focusing on regions with high diffusive DoP. The weights are chosen such that in the first phase of the minimization, the diffusive loss drives the estimation of the index of refraction $\mu$ which later helps to better disentangle material properties that occur solely in the specular component of the Mueller matrix.
Note that in our simulation, only the scalar amplitude, denoted by $|\mathbf{D}^s|$  and $|\mathbf{D}^d|$, of the depolarization matrices vary and are subsequently optimized for. Fig.~\ref{fig:baselines} validates that the proposed approach is able to successfully recover the material properties of different objects and surfaces.
As we do not optimize the surface normals, this further validates the quality of the reconstructed normals as recovering material properties without accurate normals is infeasible.

\begin{table}[t]
\small
\centering
\renewcommand{\arraystretch}{1.2}
\resizebox{1.0\linewidth}{!}{
\begin{tabular}{l@{\hspace{5mm}}c@{\hspace{3mm}}c@{\hspace{3mm}}c@{\hspace{5mm}}c@{\hspace{3mm}}c@{\hspace{3mm}}c}
\toprule[1pt]
\small{Method}&\multicolumn{3}{c}{Angular Error [\textdegree] $\downarrow$ } & \multicolumn{3}{c}{Accuracy [\%]  $\uparrow$}  \\ 
 & \small{Mean} & \small{Median}  & \small{RMSE}& \small{$3.0^{\circ}$} & \small{$5.0^{\circ}$} & \small{$10.0^{\circ}$} \\
 
\midrule[0.6pt]

SfP-DoP~\cite{atkinson2006recovery} &   65.92 & 63.76 & 70.13 & 0.02 & 0.05 & 0.24 \\

Baek et al.~\cite{Baek:PolarToF:2022} & 37.04 & 13.97 & 57.25 & 17.51 & 27.93 & 43.36 \\
PCA~\cite{Zhou2018open3d} & 18.75 & 4.86 & 35.73 & 41.18 & 52.87 & 63.74  \\
Proposed      & \textbf{15.76} & \textbf{4.27} & \textbf{28.73} & \textbf{45.94} & \textbf{55.42} & \textbf{63.80}      \\ 
\bottomrule[1pt]
\end{tabular}
}
\vspace{-2mm}
\caption{ \textbf{Quantitative evaluation for normals on experimental data.} For normal reconstruction with the real device, comparable trends to synthetic data are observable. Due to noisier ground truth and sensor imperfections, the overall error is slightly larger. However, the proposed method recovers accurate normals on real experimental data, outperforming all baseline approaches. }
\label{table:baseline_cmp}
\vspace{-2em}

\end{table}

\subsection{Experimental Evaluation}
Next, we evaluate the proposed approach on real-world data.
We pair the PolLidar sensor with a Velodyne VLS-128 reference lidar.
Fig.~\ref{fig:dataset} shows PolLidar data with ground truth distance and normals. In total, we capture 60 frames with 3 biases each and scene-adjusted laser power paired with ground-truth distance and normal information. 
For ground truth, we accumulate point clouds from the reference lidar, generate dense lidar maps, and extract normals from the meshed lidar map.

Fig.~\ref{fig:baselines_real} reports qualitative reconstruction results. Similar to the synthetic evaluations, PCA~\cite{Zhou2018open3d} introduces artifacts, whereas the proposed approach is able to recover the surface geometry correctly, e.g., the first row of Fig.~\ref{fig:baselines_real} for the transition area between ground and metal ramp. 
Furthermore, the proposed approach is able to reconstruct normals in sparse regions, e.g. for the metal support structure of the roof in the second row.
These findings are consistent with the quantitative results in Tab.~\ref{table:baseline_cmp}, where the proposed approach outperforms the best baseline by 16\% mean angular error.
For autonomous driving, accurate normals allow us to distinguish obstacles from the road and are crucial for determining if areas of the road can be overridden, e.g., detecting lost-cargo objects on roads \cite{ramos2017detecting, pinggera2016lost}. 
We show such a scenario in the last row of Fig.~\ref{fig:baselines_real}, where normals of a roadblock in 50m distance are predicted correctly as facing towards the vehicle by the proposed approach. In contrast, PCA~\cite{Zhou2018open3d} estimates the roadblock as flat with downward pointing normals likely misclassifying the object as traversable.

For distance estimation, the mean absolute error of conventional argmax-peak-finding amounts to 24 cm, whereas our method yields a mean absolute error of 20 cm outperforming the conventional distance estimation by 17\%.

\subsection{Ablation Experiments}

We further provide an ablation study in Tab.~\ref{table:ablation_all}. First, the impact of polarization cues is studied. In particular, we remove the polarization information by replacing the raw wavefronts  $\mathbf{\tilde{I}}$ with the mean over the different $\theta_i$. Removing the polarization cues, increases mean angular error by 22\%.
Furthermore, we ablate the ellipsometric reconstruction. Specifically, we remove the Mueller matrix from the inputs.
As the network needs to learn to disentangle the polarization optics of the emitter and receiver from the scene, the mean angular error of surface normal increases by 12\%.
Finally, we analyze the impact of using raw wavefronts by setting the window size to 1. Tab.~\ref{table:ablation_all} shows that the wavefront carries crucial information for scene reconstruction.

\begin{table}[t]
\begin{center}
\resizebox{1.0\linewidth}{!}{
\begin{tabular}{c@{\hspace{2mm}}c@{\hspace{2mm}}c@{\hspace{5mm}}c@{\hspace{5mm}}c@{\hspace{5mm}}c@{\hspace{5mm}}c@{\hspace{5mm}}c}
\toprule[1pt]
\multicolumn{3}{c}{\small  Ablated modules} &\small  Mean angular error [$^\circ$]\\
\small Wavefront &\small  {Polarization}&\small {Mueller} &\small    \\ 
\midrule[0.6pt]
\small  \checkmark &&&\small  10.64  \\
\small \checkmark& \small \checkmark& &\small 9.73 \\
\small  &\small \checkmark &\small \checkmark&  \small 9.47  \\
\small \checkmark  & \small \checkmark    &\small \checkmark&\small  \textbf{8.71}  \\ 
\bottomrule[1pt]
\end{tabular}
}
\end{center}
\vspace{-4mm}
\caption{ \textbf{Ablation studies for different modules on synthetic data.} The quality of the proposed method degrades when the polarization information, Mueller matrix, or wavefront is withheld. }

\label{table:ablation_all}
\vspace{-1em}
\end{table}

\section{Conclusion}
This paper introduces a novel long-range polarization wavefront lidar sensor that measures time-resolved polarization-modulated wavefronts. To recover high-resolution scene information from these raw polarimetric wavefronts, we devise a learning-based approach to recover distance, surface normals, and material properties. To train and evaluate the method, we introduce a large synthetic dataset and a real-world long-range dataset with paired raw lidar data, ground truth depth and normal maps. We validate that the proposed method improves normal and depth reconstruction by 53\% and 41\% in mean angular error and mean absolute distance error compared to existing shape-from-polarization (SfP) and ToF methods. Confirming the potential of the proposed polarimetric wavefront sensing method with a sequential acquisition setup, future work may devise parallelized acquisition setups that capture a subset of polarization states, allowing for real-time polarimetric lidar captures.

\paragraph{Acknowledgements} This work was supported by the AI-
SEE project with funding from the FFG, BMBF, and NRC-IRA.
Chenyang Lei was supported by the InnoHK program. Seung-Hwan Baek was supported by Korea NRF grant (RS-2023-00211658, 2022R1A6A1A03052954).
Felix Heide was supported by an NSF CAREER Award (2047359), a Packard Foundation Fellowship, a Sloan Research Fellowship, a Sony Young Faculty Award, a Project X Innovation Award, and an Amazon Science Research Award.

{
    \small
    \bibliographystyle{ieeenat_fullname}
    \bibliography{main}

\begin{thebibliography}{62}
\providecommand{\natexlab}[1]{#1}
\providecommand{\url}[1]{\texttt{#1}}
\expandafter\ifx\csname urlstyle\endcsname\relax
  \providecommand{\doi}[1]{doi: #1}\else
  \providecommand{\doi}{doi: \begingroup \urlstyle{rm}\Url}\fi

\bibitem[Atkinson(2017)]{atkinson2017polarisation_photometric}
Gary~A. Atkinson.
\newblock Polarisation photometric stereo.
\newblock \emph{Comput. Vis. Image Underst.}, 160:\penalty0 158--167, 2017.

\bibitem[Atkinson and Hancock(2006)]{atkinson2006recovery}
Gary~A. Atkinson and Edwin~R. Hancock.
\newblock Recovery of surface orientation from diffuse polarization.
\newblock \emph{{IEEE} Trans. Image Process.}, 15\penalty0 (6):\penalty0 1653--1664, 2006.

\bibitem[Ba et~al.(2020)Ba, Gilbert, Wang, Yang, Chen, Wang, Yan, Shi, and Kadambi]{ba2020deep}
Yunhao Ba, Alex Gilbert, Franklin Wang, Jinfa Yang, Rui Chen, Yiqin Wang, Lei Yan, Boxin Shi, and Achuta Kadambi.
\newblock Deep shape from polarization.
\newblock In \emph{ECCV}, 2020.

\bibitem[Baek et~al.(2018)Baek, Jeon, Tong, and Kim]{baek2018simultaneous}
Seung{-}Hwan Baek, Daniel~S. Jeon, Xin Tong, and Min~H. Kim.
\newblock Simultaneous acquisition of polarimetric {SVBRDF} and normals.
\newblock \emph{{ACM} Trans. Graph.}, 37\penalty0 (6):\penalty0 268:1--268:15, 2018.

\bibitem[Baek and Heide(2021)]{Baek:PolarLT:2021}
Seung-Hwan Baek and Felix Heide.
\newblock Polarimetric spatio-temporal light transport probing.
\newblock \emph{ACM Transactions on Graphics (Proc. SIGGRAPH Asia)}, 40\penalty0 (6), 2021.

\bibitem[Baek and Heide(2022)]{Baek:PolarToF:2022}
Seung-Hwan Baek and Felix Heide.
\newblock All-photon polarimetric time-of-flight imaging.
\newblock \emph{Proceedings of the IEEE Conference on Computer Vision and Pattern Recognition (CVPR)}, 2022.

\bibitem[Baek et~al.(2020)Baek, Zeltner, Ku, Hwang, Tong, Jakob, and Kim]{baek2020image}
Seung-Hwan Baek, Tizian Zeltner, Hyunjin Ku, Inseung Hwang, Xin Tong, Wenzel Jakob, and Min~H Kim.
\newblock Image-based acquisition and modeling of polarimetric reflectance.
\newblock \emph{ACM Trans. Graph.}, 39\penalty0 (4):\penalty0 139, 2020.

\bibitem[Behley et~al.(2021)Behley, Garbade, Milioto, Quenzel, Behnke, Gall, and Stachniss]{behley2021towards}
Jens Behley, Martin Garbade, Andres Milioto, Jan Quenzel, Sven Behnke, J{\"u}rgen Gall, and Cyrill Stachniss.
\newblock Towards 3d lidar-based semantic scene understanding of 3d point cloud sequences: The semantickitti dataset.
\newblock \emph{The International Journal of Robotics Research}, 40\penalty0 (8-9):\penalty0 959--967, 2021.

\bibitem[Behroozpour et~al.(2017)Behroozpour, Sandborn, Wu, and Boser]{behroozpour2017lidar}
Behnam Behroozpour, Phillip~AM Sandborn, Ming~C Wu, and Bernhard~E Boser.
\newblock Lidar system architectures and circuits.
\newblock \emph{IEEE Communications Magazine}, 55\penalty0 (10):\penalty0 135--142, 2017.

\bibitem[Berger et~al.(2017)Berger, Voorhies, and Matthies]{berger2017depth}
Kai Berger, Randolph Voorhies, and Larry~H. Matthies.
\newblock Depth from stereo polarization in specular scenes for urban robotics.
\newblock In \emph{ICRA}, 2017.

\bibitem[Carballo et~al.(2020)Carballo, Lambert, Monrroy, Wong, Narksri, Kitsukawa, Takeuchi, Kato, and Takeda]{carballo2020libre}
Alexander Carballo, Jacob Lambert, Abraham Monrroy, David Wong, Patiphon Narksri, Yuki Kitsukawa, Eijiro Takeuchi, Shinpei Kato, and Kazuya Takeda.
\newblock Libre: The multiple 3d lidar dataset.
\newblock In \emph{2020 IEEE Intelligent Vehicles Symposium (IV)}, pages 1094--1101. IEEE, 2020.

\bibitem[Carlevaris-Bianco et~al.(2016)Carlevaris-Bianco, Ushani, and Eustice]{carlevaris2016university}
Nicholas Carlevaris-Bianco, Arash~K Ushani, and Ryan~M Eustice.
\newblock University of michigan north campus long-term vision and lidar dataset.
\newblock \emph{The International Journal of Robotics Research}, 35\penalty0 (9):\penalty0 1023--1035, 2016.

\bibitem[Collett(2005)]{collett2005field}
Edward Collett.
\newblock Field guide to polarization.
\newblock Spie Bellingham, WA, 2005.

\bibitem[Cui et~al.(2017)Cui, Gu, Shi, Tan, and Kautz]{cui2017polarimetric}
Zhaopeng Cui, Jinwei Gu, Boxin Shi, Ping Tan, and Jan Kautz.
\newblock Polarimetric multi-view stereo.
\newblock In \emph{CVPR}, 2017.

\bibitem[Dave et~al.(2022)Dave, Zhao, and Veeraraghavan]{dave2022pandora}
Akshat Dave, Yongyi Zhao, and Ashok Veeraraghavan.
\newblock Pandora: Polarization-aided neural decomposition of radiance.
\newblock In \emph{European Conference on Computer Vision}, pages 538--556. Springer, 2022.

\bibitem[Deschaintre et~al.(2021)Deschaintre, Lin, and Ghosh]{deschaintre2021deep}
Valentin Deschaintre, Yiming Lin, and Abhijeet Ghosh.
\newblock Deep polarization imaging for 3d shape and svbrdf acquisition.
\newblock In \emph{CVPR}, 2021.

\bibitem[Dewan et~al.(2016)Dewan, Caselitz, Tipaldi, and Burgard]{tracking}
Ayush Dewan, Tim Caselitz, Gian~Diego Tipaldi, and Wolfram Burgard.
\newblock Motion-based detection and tracking in {3D} {L}i{DAR} scans.
\newblock In \emph{IEEE International Conference on Robotics and Automation (ICRA)}, 2016.

\bibitem[Ding et~al.(2021)Ding, Ji, Zhou, Kang, and Ye]{ding2021polarimetric}
Yuqi Ding, Yu Ji, Mingyuan Zhou, Sing~Bing Kang, and Jinwei Ye.
\newblock Polarimetric helmholtz stereopsis.
\newblock In \emph{CVPR}, 2021.

\bibitem[Dosovitskiy et~al.(2017)Dosovitskiy, Ros, Codevilla, Lopez, and Koltun]{dosovitskiy2017carla}
Alexey Dosovitskiy, German Ros, Felipe Codevilla, Antonio Lopez, and Vladlen Koltun.
\newblock Carla: An open urban driving simulator.
\newblock In \emph{Conference on robot learning}, pages 1--16. PMLR, 2017.

\bibitem[Dubayah and Drake(2000)]{dubayah2000lidar}
Ralph~O Dubayah and Jason~B Drake.
\newblock Lidar remote sensing for forestry.
\newblock \emph{Journal of forestry}, 98\penalty0 (6):\penalty0 44--46, 2000.

\bibitem[Fang et~al.(2014)Fang, Xia, Huo, and Chen]{fang2014image}
Shuai Fang, XiuShan Xia, Xing Huo, and ChangWen Chen.
\newblock Image dehazing using polarization effects of objects and airlight.
\newblock \emph{Optics express}, 22\penalty0 (16):\penalty0 19523--19537, 2014.

\bibitem[Fu et~al.(2023)Fu, Yang, Si, Zhang, Zhang, Luo, Zhan, and Zhang]{fuPolarizationSeaFog2023}
Qiang Fu, Wei Yang, Linlin Si, Meng Zhang, Yue Zhang, Kaiming Luo, Juntong Zhan, and Su Zhang.
\newblock Study of multispectral polarization imaging in sea fog environment.
\newblock \emph{Frontiers in Physics}, 11, 2023.

\bibitem[Fukao et~al.(2021)Fukao, Kawahara, Nobuhara, and Nishino]{fukao2021polarimetric}
Yoshiki Fukao, Ryo Kawahara, Shohei Nobuhara, and Ko Nishino.
\newblock Polarimetric normal stereo.
\newblock In \emph{CVPR}, 2021.

\bibitem[Geiger et~al.(2013)Geiger, Lauer, Wojek, Stiller, and Urtasun]{geiger20133d}
Andreas Geiger, Martin Lauer, Christian Wojek, Christoph Stiller, and Raquel Urtasun.
\newblock 3d traffic scene understanding from movable platforms.
\newblock \emph{IEEE transactions on pattern analysis and machine intelligence}, 36\penalty0 (5):\penalty0 1012--1025, 2013.

\bibitem[Goldstein(2011)]{Goldstein:PolarizedLight}
Dennis Goldstein.
\newblock \emph{Polarized Light}.
\newblock CRC Press, 3rd edition edition, 2011.

\bibitem[Goudreault et~al.(2023)Goudreault, Scheuble, Bijelic, Robidoux, and Heide]{goudreault2023lidarhyperoptim}
Felix Goudreault, Dominik Scheuble, Mario Bijelic, Nicolas Robidoux, and Felix Heide.
\newblock Lidar-in-the-loop hyperparameter optimization.
\newblock In \emph{CVPR}, 2023.

\bibitem[Huang et~al.(2023)Huang, Li, He, Sui, Li, and Liu]{huang2023learning}
Tianyu Huang, Haoang Li, Kejing He, Congying Sui, Bin Li, and Yun-Hui Liu.
\newblock Learning accurate 3d shape based on stereo polarimetric imaging.
\newblock In \emph{Proceedings of the IEEE/CVF Conference on Computer Vision and Pattern Recognition}, pages 17287--17296, 2023.

\bibitem[Huffman et~al.(2020)Huffman, Perring, Savage, Clot, Crouzy, Tummon, Shoshanim, Damit, Schneider, Sivaprakasam, et~al.]{huffman2020real}
J~Alex Huffman, Anne~E Perring, Nicole~J Savage, Bernard Clot, Beno{\^\i}t Crouzy, Fiona Tummon, Ofir Shoshanim, Brian Damit, Johannes Schneider, Vasanthi Sivaprakasam, et~al.
\newblock Real-time sensing of bioaerosols: Review and current perspectives.
\newblock \emph{Aerosol Science and Technology}, 54\penalty0 (5):\penalty0 465--495, 2020.

\bibitem[Huynh et~al.(2010)Huynh, Robles{-}Kelly, and Hancock]{huynh2010shape}
Cong~Phuoc Huynh, Antonio Robles{-}Kelly, and Edwin~R. Hancock.
\newblock Shape and refractive index recovery from single-view polarisation images.
\newblock In \emph{CVPR}, 2010.

\bibitem[Ikemura et~al.(2024)Ikemura, Huang, Heide, Zhang, Chen, and Lei]{ikemura2024robust}
Kei Ikemura, Yiming Huang, Felix Heide, Zhaoxiang Zhang, Qifeng Chen, and Chenyang Lei.
\newblock Robust depth enhancement via polarization prompt fusion tuning.
\newblock \emph{arXiv preprint arXiv:2404.04318}, 2024.

\bibitem[Jeon et~al.(2023)Jeon, Meuleman, Baek, and Kim]{jeon2023polarimetric}
Daniel~S Jeon, Andr{\'e}as Meuleman, Seung-Hwan Baek, and Min~H Kim.
\newblock Polarimetric itof: Measuring high-fidelity depth through scattering media.
\newblock In \emph{Proceedings of the IEEE/CVF Conference on Computer Vision and Pattern Recognition}, pages 12353--12362, 2023.

\bibitem[Kadambi et~al.(2015)Kadambi, Taamazyan, Shi, and Raskar]{kadambi2015polarized}
Achuta Kadambi, Vage Taamazyan, Boxin Shi, and Ramesh Raskar.
\newblock Polarized 3d: High-quality depth sensing with polarization cues.
\newblock In \emph{ICCV}, 2015.

\bibitem[Kellner et~al.(2019)Kellner, Armston, Birrer, Cushman, Duncanson, Eck, Falleger, Imbach, Kr\'{a}l, Kr\u{u}{\v{c}}ek, et~al.]{kellner2019new}
James~R Kellner, John Armston, Markus Birrer, KC Cushman, Laura Duncanson, Christoph Eck, Christoph Falleger, Benedikt Imbach, Kamil Kr\'{a}l, Martin Kr\u{u}{\v{c}}ek, et~al.
\newblock New opportunities for forest remote sensing through ultra-high-density drone lidar.
\newblock \emph{Surveys in Geophysics}, 40:\penalty0 959--977, 2019.

\bibitem[Kondo et~al.(2020)Kondo, Ono, Sun, Hirasawa, and Murayama]{kondo2020accurate}
Yuhi Kondo, Taishi Ono, Legong Sun, Yasutaka Hirasawa, and Jun Murayama.
\newblock Accurate polarimetric {BRDF} for real polarization scene rendering.
\newblock In \emph{ECCV}, 2020.

\bibitem[Lei et~al.(2020)Lei, Huang, Zhang, Yan, Sun, and Chen]{lei2020polarized}
Chenyang Lei, Xuhua Huang, Mengdi Zhang, Qiong Yan, Wenxiu Sun, and Qifeng Chen.
\newblock Polarized reflection removal with perfect alignment in the wild.
\newblock In \emph{CVPR}, 2020.

\bibitem[Lei et~al.(2022)Lei, Qi, Xie, Fan, Koltun, and Chen]{lei2022shape}
Chenyang Lei, Chenyang Qi, Jiaxin Xie, Na Fan, Vladlen Koltun, and Qifeng Chen.
\newblock Shape from polarization for complex scenes in the wild.
\newblock In \emph{Proceedings of the IEEE/CVF Conference on Computer Vision and Pattern Recognition}, pages 12632--12641, 2022.

\bibitem[Linnhoff et~al.(2022)Linnhoff, Scheuble, Bijelic, Elster, Rosenberger, Ritter, Dai, and Winner]{linnhoff2022simulating}
Clemens Linnhoff, Dominik Scheuble, Mario Bijelic, Lukas Elster, Philipp Rosenberger, Werner Ritter, Dengxin Dai, and Hermann Winner.
\newblock Simulating road spray effects in automotive lidar sensor models.
\newblock \emph{arXiv preprint arXiv:2212.08558}, 2022.

\bibitem[Lyu et~al.(2019)Lyu, Cui, Li, Pollefeys, and Shi]{lyu2019reflection}
Youwei Lyu, Zhaopeng Cui, Si Li, Marc Pollefeys, and Boxin Shi.
\newblock Reflection separation using a pair of unpolarized and polarized images.
\newblock In \emph{NeurIPS}, 2019.

\bibitem[Miyazaki et~al.(2003)Miyazaki, Tan, Hara, and Ikeuchi]{miyazaki2003polarization}
Daisuke Miyazaki, Robby~T. Tan, Kenji Hara, and Katsushi Ikeuchi.
\newblock Polarization-based inverse rendering from a single view.
\newblock In \emph{ICCV}, 2003.

\bibitem[Miyazaki et~al.(2016)Miyazaki, Shigetomi, Baba, Furukawa, Hiura, and Asada]{Miyazaki2016SurfaceNE}
Daisuke Miyazaki, Takuya Shigetomi, Masashi Baba, Ryo Furukawa, Shinsaku Hiura, and Naoki Asada.
\newblock Surface normal estimation of black specular objects from multiview polarization images.
\newblock \emph{Optical Engineering}, 56\penalty0 (4):\penalty0 041303, 2016.

\bibitem[Pinggera et~al.(2016)Pinggera, Ramos, Gehrig, Franke, Rother, and Mester]{pinggera2016lost}
Peter Pinggera, Sebastian Ramos, Stefan Gehrig, Uwe Franke, Carsten Rother, and Rudolf Mester.
\newblock Lost and found: detecting small road hazards for self-driving vehicles.
\newblock In \emph{2016 IEEE/RSJ International Conference on Intelligent Robots and Systems (IROS)}, pages 1099--1106. IEEE, 2016.

\bibitem[Polyanskiy()]{iordatabase}
Mikhail~N. Polyanskiy.
\newblock Refractive index database.
\newblock \url{https://refractiveindex.info}.
\newblock Accessed on 2023-11-12.

\bibitem[Rahmann and Canterakis(2001)]{rahmann2001reconstruction}
Stefan Rahmann and Nikos Canterakis.
\newblock Reconstruction of specular surfaces using polarization imaging.
\newblock In \emph{CVPR}, 2001.

\bibitem[Ramos et~al.(2017)Ramos, Gehrig, Pinggera, Franke, and Rother]{ramos2017detecting}
Sebastian Ramos, Stefan Gehrig, Peter Pinggera, Uwe Franke, and Carsten Rother.
\newblock Detecting unexpected obstacles for self-driving cars: Fusing deep learning and geometric modeling.
\newblock In \emph{2017 IEEE Intelligent Vehicles Symposium (IV)}, pages 1025--1032. IEEE, 2017.

\bibitem[Risb{\o}l and Gustavsen(2018)]{risbol2018lidar}
Ole Risb{\o}l and Lars Gustavsen.
\newblock Lidar from drones employed for mapping archaeology--potential, benefits and challenges.
\newblock \emph{Archaeological Prospection}, 25\penalty0 (4):\penalty0 329--338, 2018.

\bibitem[Riviere et~al.(2017)Riviere, Reshetouski, Filipi, and Ghosh]{riviere2017polarization}
J{\'{e}}r{\'{e}}my Riviere, Ilya Reshetouski, Luka Filipi, and Abhijeet Ghosh.
\newblock Polarization imaging reflectometry in the wild.
\newblock \emph{{ACM} Trans. Graph.}, 36\penalty0 (6):\penalty0 206:1--206:14, 2017.

\bibitem[Royo and Gras(United States Patent 9689667B2 2013)]{royo2012lightbeam}
Santiago~Royo Royo and Jordi~Riu Gras.
\newblock System, method and computer program for receiving a light beam, United States Patent 9689667B2 2013.

\bibitem[Sassen(1991)]{sassen1991polarization}
Kenneth Sassen.
\newblock The polarization lidar technique for cloud research: A review and current assessment.
\newblock \emph{Bulletin of the American Meteorological Society}, 72\penalty0 (12):\penalty0 1848--1866, 1991.

\bibitem[Sassen(2005)]{sassen2005polarization}
Kenneth Sassen.
\newblock Polarization in lidar.
\newblock In \emph{LIDAR: Range-resolved optical remote sensing of the atmosphere}, pages 19--42. Springer, 2005.

\bibitem[Schechner et~al.(2001)Schechner, Narasimhan, and Nayar]{schechner2001instant}
Yoav~Y. Schechner, Srinivasa~G. Narasimhan, and Shree~K. Nayar.
\newblock Instant dehazing of images using polarization.
\newblock In \emph{CVPR}, 2001.

\bibitem[Schotland et~al.(1971)Schotland, Sassen, and Stone]{schotland1971observations}
Richard~M Schotland, Kenneth Sassen, and Richard Stone.
\newblock Observations by lidar of linear depolarization ratios for hydrometeors.
\newblock \emph{Journal of Applied Meteorology and Climatology}, 10\penalty0 (5):\penalty0 1011--1017, 1971.

\bibitem[Shakeri et~al.(2021)Shakeri, Loo, Zhang, and Hu]{shakeri2021polarimetric}
Moein Shakeri, Shing~Yang Loo, Hong Zhang, and Kangkang Hu.
\newblock Polarimetric monocular dense mapping using relative deep depth prior.
\newblock \emph{IEEE Robotics and Automation Letters}, 6\penalty0 (3):\penalty0 4512--4519, 2021.

\bibitem[Shi et~al.(2020)Shi, Guo, Jiang, Wang, Shi, Wang, and Li]{PV-RCNN}
Shaoshuai Shi, Chaoxu Guo, Li Jiang, Zhe Wang, Jianping Shi, Xiaogang Wang, and Hongsheng Li.
\newblock {PV-RCNN}: Point-voxel feature set abstraction for {3D} object detection.
\newblock In \emph{IEEE/CVF Conference on Computer Vision and Pattern Recognition (CVPR)}, 2020.

\bibitem[Tian et~al.(2023)Tian, Pan, Wang, Mao, Zhang, Bao, Tan, and Cui]{tian2023dps}
Chaoran Tian, Weihong Pan, Zimo Wang, Mao Mao, Guofeng Zhang, Hujun Bao, Ping Tan, and Zhaopeng Cui.
\newblock Dps-net: Deep polarimetric stereo depth estimation.
\newblock In \emph{Proceedings of the IEEE/CVF International Conference on Computer Vision}, pages 3569--3579, 2023.

\bibitem[Tian et~al.(2021)Tian, Zheng, Zou, Li, and Zhang]{tian2021dynamic}
Shishun Tian, Minghuo Zheng, Wenbin Zou, Xia Li, and Lu Zhang.
\newblock Dynamic crosswalk scene understanding for the visually impaired.
\newblock \emph{IEEE transactions on neural systems and rehabilitation engineering}, 29:\penalty0 1478--1486, 2021.

\bibitem[Treibitz and Schechner(2009)]{treibitz2009active}
Tali Treibitz and Yoav~Y. Schechner.
\newblock Active polarization descattering.
\newblock \emph{{IEEE} Trans. Pattern Anal. Mach. Intell.}, 31\penalty0 (3):\penalty0 385--399, 2009.

\bibitem[Vasilkov et~al.(2001)Vasilkov, Goldin, Gureev, Hoge, Swift, and Wright]{vasilkov2001airborne}
Alexander~P Vasilkov, Yury~A Goldin, Boris~A Gureev, Frank~E Hoge, Robert~N Swift, and C~Wayne Wright.
\newblock Airborne polarized lidar detection of scattering layers in the ocean.
\newblock \emph{Applied Optics}, 40\penalty0 (24):\penalty0 4353--4364, 2001.

\bibitem[Weitkamp(2006)]{weitkamp2006lidar}
Claus Weitkamp.
\newblock \emph{Lidar: range-resolved optical remote sensing of the atmosphere}.
\newblock Springer Science \& Business, 2006.

\bibitem[Yoshida et~al.(2018)Yoshida, Golyanik, Wasenm{\"u}ller, and Stricker]{yoshida2018improving}
Tomonari Yoshida, Vladislav Golyanik, Oliver Wasenm{\"u}ller, and Didier Stricker.
\newblock Improving time-of-flight sensor for specular surfaces with shape from polarization.
\newblock In \emph{2018 25th IEEE International Conference on Image Processing (ICIP)}, pages 1558--1562. IEEE, 2018.

\bibitem[Zhang and Singh(2014)]{ZhangLOAM2014}
Ji Zhang and Sanjiv Singh.
\newblock {LOAM} : {L}i{DAR} odometry and mapping in real-time.
\newblock \emph{Robotics: Science and Systems Conference (RSS)}, 2014.

\bibitem[Zhou et~al.(2018)Zhou, Park, and Koltun]{Zhou2018open3d}
Qian{-}Yi Zhou, Jaesik Park, and Vladlen Koltun.
\newblock {Open3D}: {A} modern library for {3D} data processing.
\newblock \emph{CoRR}, abs/1801.09847, 2018.

\bibitem[Zhu and Smith(2019)]{zhu2019depth}
Dizhong Zhu and William A.~P. Smith.
\newblock Depth from a polarisation + {RGB} stereo pair.
\newblock In \emph{CVPR}, 2019.

\end{thebibliography}
}

\clearpage


\section*{Supplementary material}

# Polarization Wavefront Lidar:
# Learning Large Scene Reconstruction from Polarized Wavefronts
# (Supplementary Information)

Dominik Scheuble[1,2]* Chenyang Lei [5]* Seung-Hwan Baek[4] Mario Bijelic[3,5] Felix Heide[3,5]

[1]Mercedes-Benz AG  [2]TU Darmstadt  [3]Torc Robotics  [4]POSTECH  [5]Princeton University

In this supplemental document, we present additional details on the PolLidar prototype, the simulation and reconstruction method, the testing and training datasets, and the training process. We also provide additional quantitative and qualitative results in support of the findings from the main manuscript.

## Contents

## 1. PolLidar Prototype

As the PolLidar is an entirely novel prototype, we provide additional details about its construction (Sec. 1.1); subsequently, we share insight into key characteristics of the sensor (Sec. 1.2); then, we validate that the surface-induced polarization cues can be captured with the proposed prototype (Sec. 1.3); finally, we discuss improvements that will be made for future prototypes (Sec. 1.4).

---

*These authors contributed equally to this work.

## 1.1. Prototype Construction

The PolLidar extends the concept from Beamagines L3CAM lidar [1] starting from traditional ToF and adding polarization capabilities. However, the emitter and receiver are a redesign with custom opto-mechanics to fit waveplates and polarizer necessary to modulate the polarization. Fig. 1 illustrates a sectioned view of emitter and receiver to showcase the polarization optics. All used parts are listed in Tab. 1. The optics are designed to minimize the angle of incidence (AoI) onto the waveplates and polarizer. The maximum AoI on the emitter side is 2°, whereas on the receiving side, the maximum AoI is 8°. The optics on the emitting side allow for beam divergence of 0.364°. On the receiving side, an entrance pupil of 6.8 mm is realized. An overview of sensor key specifications is provided by Tab. 2.

The raw wavefront is captured with a National Instruments PCIe-5764 FlexRIO-Digitizer ADC, sampling at 1 Gs/s. The sampling frequency allows for 1 ns wide bins resulting in a 15cm range resolution. The ADC is interfaced from a LabView application which triggers the acquisition after the piezoelectric motors are finished rotating the waveplates or linear polarizer to their respective rotation angle $\theta_i$. Raw wavefronts for a single frame with 150 rows and 236 columns amount to approximately 100 MB of raw binary data where two bytes per bin are used to encode the measured intensity. For the acquisition, a Python API interfacing the LabView application has been developed that can be leveraged in future works for e.g. online optimization as discussed in [6].

| Item# | Part Description | Quantity | Model Name |
|---|---|---|---|
| 1 | Silicon-based Avalanche Photo Diode (APD) | 1 | Proprietary |
| 2 | Pulsed Laser source | 1 | Proprietary |
| 3 | Polarization-maintaining (PM) Fiber $10.5\,\mu m$ | 1 | Coherent PM1060L |
| 4 | Emitter fiber Coliminator | 1 | Thorlabs F230APC-1064 |
| 5 | Emitter Lens 1 | 1 | Edmund Optics 67-537 |
| 6 | Emitter Lens 2 | 1 | Edmund Optics 67-998 |
| 7 | Emitter Lens 3 | 1 | Edmund Optics 68-001 |
| 8 | Receiver Lens 1 | 1 | Edmund Optics 68-001 |
| 10 | Receiver Optics | 3 | Proprietary |
| 12 | Band-pass Filter 1064nm | 1 | Proprietary |
| 13 | Quarter-wave Plate | 2 | Thorlabs WPQ10M-1064 |
| 14 | Half-wave Plate | 1 | Thorlabs WPH10M-1064 |
| 15 | Linear Polarizer | 1 | Throlabs LPNIRB100 |
| 16 | Piezoelectric Motors | 4 | Thorlabs ELL14K |
| 17 | MEMS Micro Mirror Scanning System | 1 | Proprietary |
| 18 | Micro Mirror Receiving System | 1 | Proprietary |

Table 1. List of used parts for receiver and emitter.

| Specification | Value |
|---|---|
| Vertical resolution | 150 rows over 23.95° |
| Horizontal resolution | 236 columns over 31.05° |
| Temporal resolution | 1488 bins over 223 m |
| Rotation angle $\theta_i$ resolution | 0.01 ° |
| Wavelength | 1064 nm |
| Beam divergence | 0.326° |
| Max. AoI on polarizing optics | 2°(emitter), 8°(receiver) |

Table 2. Key specification of constructed PolLidar prototype.

## 1.2. System Response and Noise Characteristics

The polarization cues necessary for Shape-from-Polarization (SfP) approaches are reflected in the intensity readings of the ADC. Depending on the degree-of-polarization (DoP), the intensity change for different polarization states can be, as function

---
[1] https://beamagine.com/product.

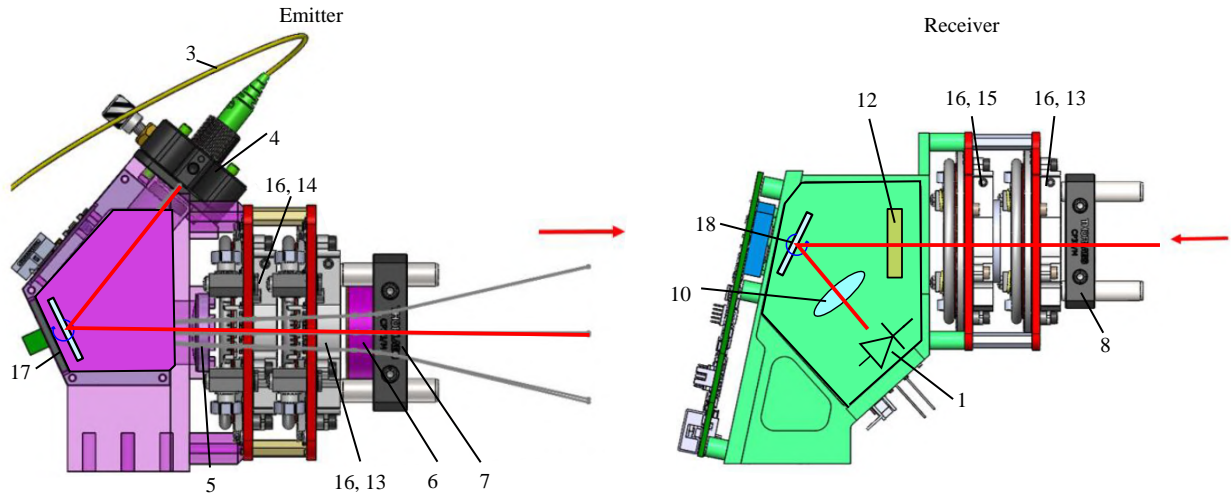

Figure 1. Sectioned view of emitter (left) and receiver (right). Parts numbers can be referenced from Tab. 1. For the protection of intellectual property, parts of the sectioned view are covered and replaced by schematic placeholders.

of geometry, material and viewing direction, very subtle. Thus, accurate intensity readings are crucial for a successful reconstruction. Hence, we provide inside in the system response of the PolLidar towards the key parameters of laser power and bias voltage of the APD. Furthermore, we investigate the noise characteristics of the sensor to better distinguish polarization-induced intensity change from sensor noise.

**System Response** To obtain meaningful intensity readings, laser power and bias need to be adjusted according to the scene. In contrast to conventional lidar sensors, we allowed access to these low-level parameters. We investigate the system response to laser power and bias in a controlled environment as shown in Fig. 2. Here, we repeatedly measure a single pixel on a calibrated 90% reflection target for different biases and laser powers. We average 50 frames per laser power/bias to limit the effect of noise. We find that large bias values quickly lead to saturation at around 0.4V. This will render the intensities meaningless for SfP as all polarization states will likely be saturated making reconstruction unfeasible. In contrast, small biases result in low intensity readings below the noise floor and subsequently dropped points preventing reconstruction altogether. As indicated by Fig. 2, we observe an exponential increase in intensity depending on the bias. Thus, the bias operating point needs to be selected with care to allow for successful reconstruction. On the other hand, we find that the laser power is well behaved in this regard and find an approximately linear relationship between laser power and intensity. Motivated by the findings in Fig. 2, we opt to select one laser power per captured frame and but always collect frames with 3 different biases. Details on acquisition are provided in Sec. 3.2.

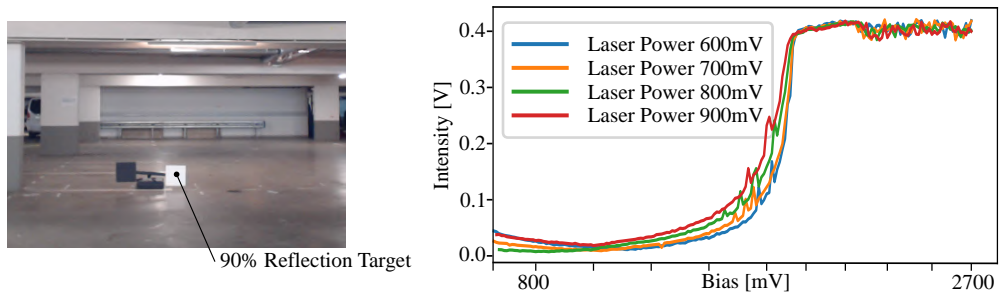

Figure 2. System response towards bias and laser power. We observe an approximately exponential relationship between bias and intensity but only a linear one between laser power and intensity for a single pixel on a 90% reflection target. We average 50 frames per laser power, bias pair to reduce the effect of noise.

**Noise Model** Furthermore, we investigate the noise characteristics of the sensor. To this end, we continuously capture the same scene 800 times as shown by Fig. 3. On the right of Fig. 3, we show intensity distributions for four selected pixels on interesting targets. We observe that the noise can be described as approximately Gaussian centered around a single mean. Thereby, we ensure to operate the laser diode and detector with sufficient warm up, such that over the measurements no temperature drift is possible. However, there is a substantial deviation of the intensity with standard deviations up to 0.03 mV which is coherent with the specifications of the installed APD. Therefore for some regions of a captured frame, the intensity deviation from noise will likely overshadow the desirable deviation observable for different polarization states. As a result, we opt to average 10 frames to increase resilience from sensor noise, as further discussed in Sec. 3.2. In addition, we test the sensor for repeatability by returning to the same rotation angles $\theta_i$ after setting a series of different rotation angles. We observe no drift in this regard.

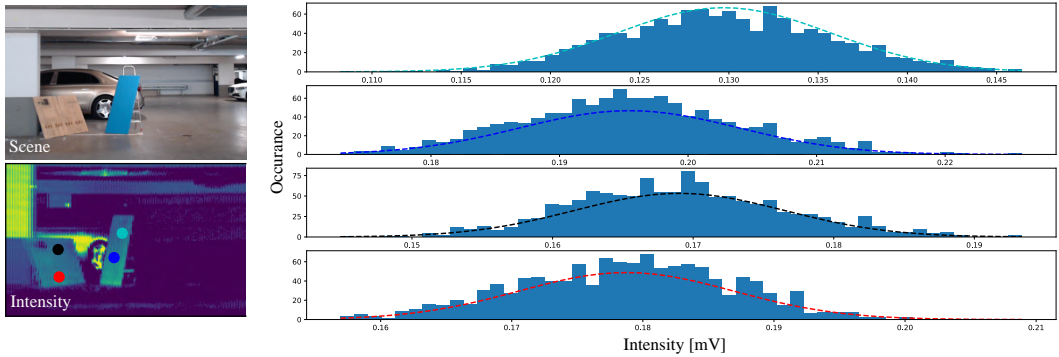

Figure 3. Noise characteristics of the PolLidar. We repeatedly measure the same scene 800 times. On the right, we show intensity distributions for selected pixels. The intensities are centered around a mean indicating no e.g. temperature-dependent drift. However, we observe substantial deviations around the mean, indicating that multiple wavefronts must be averaged to reduce the sensor noise.

### 1.3. Assessment of Polarization Cues

Our proposed SfP reconstruction method relies on rotating ellipsometry for acquisition. However, the results from rotating ellipsometry are not directly interpretable and do not allow an intuitive assessment weather the PolLidar is able to capture scene-dependent polarization cues. To this end, we follow the experiments from [1]. We fix the rotation angles of HWP and QWP $\theta^1$ and $\theta^2$ to $\theta^1 = \theta^2 = 0$. When assuming the scene to be solely diffusive, the azimuth angle of the surface normal can be directly found using the LP. Intuitively, the observed intensity $I$ after the LP will have maxima where the rotation angle $\theta^4$ of the LP and the angle of polarization of the reflected light align. In theory, the intensity varies sinusoidally with respect to $\theta^4$. The phase-shift $\phi$ or position of the maxima translates directly to the azimuth angle of the surface normal up to an ambiguity of 180°. The measured intensity $I$ at the APD is given as

$$I(\phi, \theta^4) = \frac{I_{\max} + I_{\min}}{2} + \frac{I_{\max} - I_{\min}}{2} \cos\left(2\theta^4 - 2\phi\right), \tag{1}$$

where $I_{\max}$ and $I_{\min}$ are the maximal and minimal observed intensities, respectively. Note that for our setup, the rotation angle $\theta^3$ must be set to $\theta^3 = \theta^4$ to eliminate the effect of the QWP in the receiver.

We use this acquisition approach as an intuitive assessment of the PolLidar. To this end, we setup scenes with flat targets placed with different azimuth angles in front of the sensor. An example is shown in Fig. 4, where the object in focus is the blue metal plate oriented to the right and left, respectively. Next, we increase the rotation angle $\theta^4$ of the LP in steps of 2° and average 10 wavefronts per rotation angle to limit the effect of sensor noise. As shown on the right of Fig. 4, we observe that the intensity for both scenes can be described as a sinusoid. Furthermore, the phase-shift between left- and right-oriented target is clearly visible and is close to the difference in the azimuth angle of the two metal plates. As such, we find that PolLidar is able to scan a scene accurately enough to capture polarization cues.

### 1.4. Real-Time Capability

By design, the PolLidar prototype is not real-time capable. To allow the largest possible flexibility we opted for finely controllable orientations of the wave plates, allowing us to adjust the polarization states instead and capturing all possible

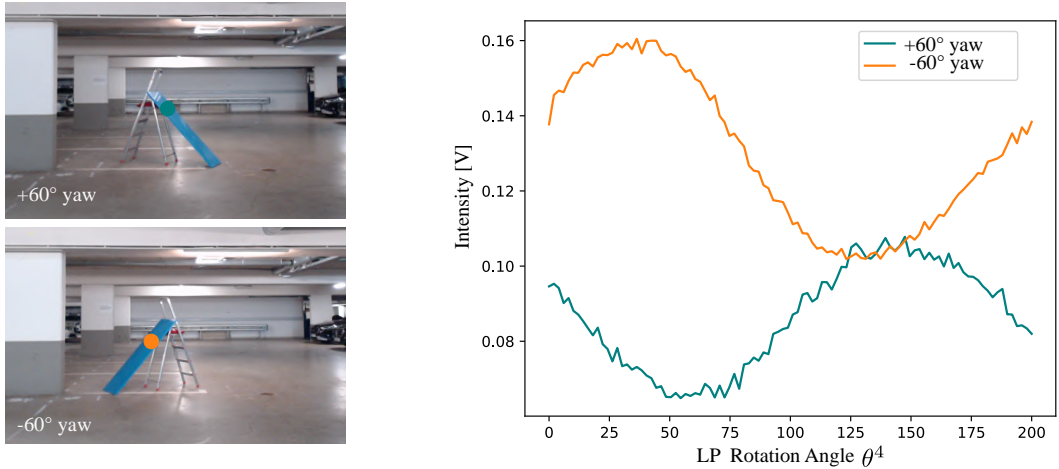

Figure 4. Assessment of Polarization Cues. We increase the rotation angle of the LP $\theta^4$ in steps of 2° and observe a single pixel on the right- and leftwards oriented blue metal plate. We find that the intensity varies sinusoidally with respect to $\theta^4$ and we observe a phase shift depending on the normal, adhering well with the theory and experiments from [1].

combinations of emitted and captured polarization states. As a consequence setting the polarization state requires considerable time. However, we imagine running the measurement in parallel as in passive cameras which employ four constant states.[2]. In detail, this would require multiple linear polarizers at fixed rotation angles, which are placed in front of an array of APDs similar to a Bayer-Pattern for RGB cameras. In summary, this work contributes a first proof showcasing polarization lidar bridging previous application domains allowing required distances and resolutions for autonmous driving vehicles especially enhancing the geometric representation for high distant objects.

Future works may optimize the number of required rotation angles for real-time capable reconstruction in automotive scenarios.

## 2. Lidar Forward Model and Simulator

In the following, we formalize the components of the polarimetric lidar forward model in Section 2.1. Then, we discuss the necessary changes made to the CARLA simulator and lastly show how we tune noise characteristics.

### 2.1. Temporal-Polarimetric Lidar Forward Model

Following the formulation from Baek et al. [2], we provide an overview of the individual components of the lidar forward model in the following.

**Linear Polarizer** A linear polarizer at a rotation angle $\theta$ with respect to a reference axis has a Mueller matrix given as

$$\mathbf{L} = \frac{1}{2} \begin{bmatrix} 1 & \cos 2\theta & \sin 2\theta & 0 \\ \cos 2\theta & \cos^2 2\theta & \cos 2\theta \sin 2\theta & 0 \\ \sin 2\theta & \cos 2\theta \sin 2\theta & \sin^2 2\theta & 0 \\ 0 & 0 & 0 & 0 \end{bmatrix}. \qquad (2)$$

---
[2]see for example the Sony IMX253MZR image, https://www.sony-semicon.com/files/62/flyer_industry/IMX250_264_253MZR_MYR_Flyer_en.pdf.

**Wave Plate** The Mueller matrix of a wave plate with retardance $\phi$ at angle $\theta$ to the horizontal is defined as

$$\mathbf{R} = \begin{bmatrix} 1 & 0 & 0 & 0 \\ 0 & \mathbf{R}_{11} & \mathbf{R}_{12} & \mathbf{R}_{13} \\ 0 & \mathbf{R}_{21} & \mathbf{R}_{22} & \mathbf{R}_{23} \\ 0 & \mathbf{R}_{31} & \mathbf{R}_{32} & \mathbf{R}_{33} \end{bmatrix},$$

$$\mathbf{R}_{11} = \cos^2 2\theta + \sin^2 2\theta \cos \phi,$$
$$\mathbf{R}_{12} = \sin 2\theta \cos 2\theta (1 - \cos \phi),$$
$$\mathbf{R}_{13} = \sin 2\theta \sin \phi,$$
$$\mathbf{R}_{21} = \sin 2\theta \cos 2\theta (1 - \cos \phi),$$
$$\mathbf{R}_{22} = \cos^2 2\theta \cos \phi + \sin^2 2\theta,$$
$$\mathbf{R}_{23} = -\cos 2\theta \sin \phi,$$
$$\mathbf{R}_{31} = -\sin 2\theta \sin \phi,$$
$$\mathbf{R}_{32} = \cos 2\theta \sin \phi,$$
$$\mathbf{R}_{33} = \cos \phi. \tag{3}$$

Ideal half/quarter-wave plates have a retardance values of $\pi$ and $\pi/2$, respectively, resulting in the Mueller matrices

$$\mathbf{W}(\theta) = \mathbf{R}(\theta, \phi = \pi), \tag{4}$$
$$\mathbf{Q}(\theta) = \mathbf{R}(\theta, \phi = \pi/2). \tag{5}$$

**Surface Fresnel Reflection and Transmission** The transmitted and reflected components are represented as the Fresnel Mueller matrices

$$\mathbf{F}_{T,R} = \begin{bmatrix} \frac{F^\perp + F^\parallel}{2} & \frac{F^\perp - F^\parallel}{2} & 0 & 0 \\ \frac{F^\perp - F^\parallel}{2} & \frac{F^\perp + F^\parallel}{2} & 0 & 0 \\ 0 & 0 & \sqrt{F^\perp F^\parallel} \cos \delta & \sqrt{F^\perp F^\parallel} \sin \delta \\ 0 & 0 & -\sqrt{F^\perp F^\parallel} \sin \delta & \sqrt{F^\perp F^\parallel} \cos \delta \end{bmatrix}. \tag{6}$$

We compute the perpendicular and parallel Fresnel coefficients $F^{\perp,\parallel}$ for reflection and transmission [4], respectively. $\delta$ is the phase shift that has the value of $\phi$ or 0 for the dielectric component.

**Coordinate Conversion** A coordinate-conversion Mueller matrix has the following form

$$\mathbf{C} = \begin{bmatrix} 1 & 0 & 0 & 0 \\ 0 & \cos 2\theta & \sin 2\theta & 0 \\ 0 & -\sin 2\theta & \cos 2\theta & 0 \\ 0 & 0 & 0 & 1 \end{bmatrix}, \tag{7}$$

where $\theta$ is the rotation angle of the input $x$ basis vector.

**Depolarization Matrices** The depolarization matrices for diffuse $\mathbf{D}^d$ and specular reflection $\mathbf{D}^s$ are diagonal matrices where the j-th diagonal entry is given by

$$\mathbf{D}_j^{\{d,s\}}(\tau) = |\mathbf{D}|^{\{d,s\}} \exp\left(\frac{(\tau - t_{\text{peak}}^2)}{2\sigma^2}\right). \tag{8}$$

As denoted in the main paper, $|\mathbf{D}^d|$ and $|\mathbf{D}^s|$ describe the amplitude of the depolarization matrices and $t_{\text{peak}}$ is the temporal index of the wavefront peak. The parameter $\sigma$ defines the pulse width which we tune such that the pulse width of downsampled synthetic rays resemble the pulse width of the real device.

**Micro-facet Distribution** We use the Smith Shading and masking term $G$ from [7] and the GGX facet distribution term $D$ from [11] defined as follows

$$D(\theta_h; m) = \frac{m^2}{\pi \cos^4 \theta_h (m^2 + \tan^2 \theta_h)^2},$$
$$G(\theta_i, \theta_o; m) = \frac{2}{1 + \sqrt{1 + m^2 \tan^2 \theta_i}} \frac{2}{1 + \sqrt{1 + m^2 \tan^2 \theta_o}}, \quad (9)$$

where $\theta_h$ is the half-way angle, $\theta_i$ and $\theta_o$ are the incident and outgoing angles. $m$ is the surface roughness.

**Lidar Forward Model** With all individual components of the lidar forward model in hand, the Mueller matrix $\mathbf{M}$ of the scene is given as sum of specular and diffuse reflection as

$$\mathbf{M}(\tau) = \underbrace{\frac{D(\theta_h; m) G(\theta_i, \theta_o; m)}{4 \cos \theta_i \cos \theta_o} \mathbf{D}^s(\tau) \mathbf{F}_R}_{\text{specular}} + \underbrace{\mathbf{C}_{\text{n} \to \omega} \mathbf{F}_T^o \mathbf{D}^d(\tau) \mathbf{F}_T^i \mathbf{C}_{\omega \to \text{n}}}_{\text{diffuse}}, \quad (10)$$

where $\theta_h = \cos^{-1}(\mathbf{h} \cdot \mathbf{n})$, $\theta_i = \cos^{-1}(\mathbf{n} \cdot \boldsymbol{\omega})$, $\theta_o = \cos^{-1}(\mathbf{n} \cdot \boldsymbol{\omega})$, $\mathbf{n}$ is the surface normal and $\boldsymbol{\omega}$ is the viewing direction. After applying distance dependent attenuation and the cosine shading, the temporal-polarimetric Mueller matrix of the scene can be written as

$$\mathbf{H} = \frac{\mathbf{n} \cdot \boldsymbol{\omega}}{d^2} \mathbf{M}. \quad (11)$$

The Mueller matrix of the emitter $\mathbf{P}_i$, consisting of HWP and QWP, is defined as

$$\mathbf{P}_i = \mathbf{Q}(\theta_i^2) \mathbf{W}(\theta_i^1). \quad (12)$$

The Mueller matrix of the receiver $\mathbf{A}_i$, consisting of QWP and LP, is defined as

$$\mathbf{A}_i = \mathbf{L}(\theta_i^4) \mathbf{Q}(\theta_i^3). \quad (13)$$

The laser outputs horizontally polarized light defined by the Stokes vector $\mathbf{s}_{\text{laser}} = [1, 1, 0, 0]^T$. Hence, the Stokes vector of the light received at the APD can be modeled as

$$\mathbf{s}_{\text{APD}} = \mathbf{P}_i(\theta_i) \mathbf{H} \mathbf{P}_i(\theta_i) \mathbf{s}_{\text{laser}} \quad (14)$$

where the first element of $\mathbf{s}_{\text{APD}}$ equals the intensity measured by the APD.

## 2.2. Implementation Details of Polarization CARLA Simulator

We implement the previously discussed polarimetric lidar forward in CARLA. To this end, we rely on the full-wavefront lidar simulator presented by [6]. We extract the necessary normals $\mathbf{n}$, index of refraction $\mu$, diffuse and specular depolarization amplitude $|\mathbf{D}^d|$, $|\mathbf{D}^s|$ and roughness $m$ from CARLA. Analogous to [6], we use custom material cameras for diffuse amplitude $|\mathbf{D}^d|$, specular amplitude $|\mathbf{D}^s|$, roughness $m$. When following the notation of [6], the diffuse amplitude equals $d$, the specular amplitude equals $s$ and roughness translates to $\alpha$.

However, the index of refraction $\mu$ is not assigned to materials in CARLA by default. To this end, we modify the CARLA ray-tracer to return a material ID for each hit point. When the ray-tracer returns a hit point, we query the face of the mesh at this hit point for the name of the assigned material. Worlds in CARLA have more than 5000 assigned materials. We cluster the materials based on their name and their respective parent material. In total, we define 10 clusters and assign an index of refraction to each cluster, which we subsequently look-up during rendering. This method is suitable for static objects like buildings, infrastructure, or vegetation. However, moving objects such as vehicles in CARLA are considered only with a simplified mesh to reduce the computational cost for the ray-tracing. This simplified mesh does not have any materials assigned to it as it was not intended for rendering but only for extracting distance information with the ray-tracer. We circumvent this problem by manually assigning materials to each face of a vehicle mesh as shown in Fig. 5.

Finally, we extend the ray-tracer to return the surface normal $\mathbf{n}$ for each hit point. The Unreal Engine underlying CARLA provides a straightforward interface for querying the mesh for normals at a hit point. For extracting the normal, we add an extra channel for the normals to the lidar model in the C++ code and adapt the auxiliary Python interface accordingly. As the Unreal Engine returns normals in world-coordinates, we transform the normals into the local sensor frame.

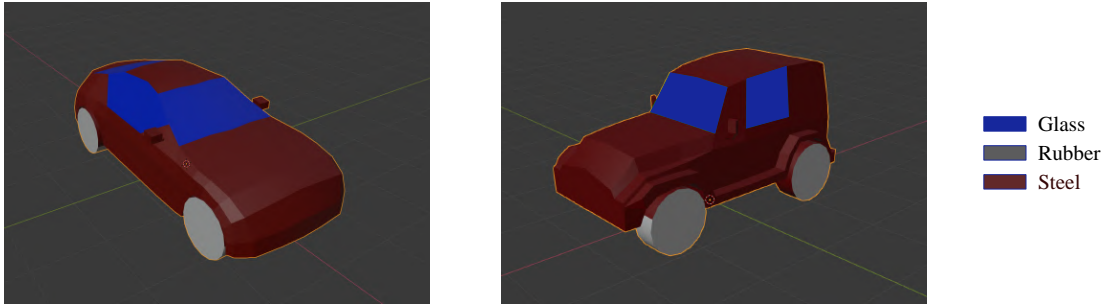

Figure 5. Assigned material to car meshes in CARLA. The ray-tracer in CARLA uses simplified meshes for cars to reduce rendering costs. These meshes do not have a material assigned to them. We manually assign 3 materials to each face of the car meshes, e.g., glass, steel and rubber.

### 2.3. Simulating Sensor Noise

The wavefronts $\tilde{\mathbf{I}}$ obtained from our Carla simulator are noise-free, and we corrupt the synthetic images by adding simulated sensor noise. The raw wavefronts are added by a noise $\eta_{\text{SENSOR}}(\tilde{\mathbf{I}})$, which consists of a Poissonian signal-dependent noise and a Gaussian signal-independent component

$$\eta_{\text{SENSOR}}(\tilde{\mathbf{I}}) = \eta_p(\tilde{\mathbf{I}}, a_p) + \eta_g(\sigma_g), \tag{15}$$

where $\eta_p$ a Poissonian signal-dependent component, and $\eta_g$ a Gaussian signal-independent component. To choose the correct parameters for $a_p$ and $\sigma_g$, we rely on the noise characteristics presented in Sec. 1.2 and set $a_p$ and $\sigma_g$ to 1e-3 and 1e-4, respectively.

## 3. Additional Detail on Dataset

In the following, we provide additional details about ground truth generation and acquisition of the real-world large-scene polarimetric lidar dataset.

### 3.1. Ground Truth

As shown in Fig. 6, for generating ground truth distance and normals, we pair the PolLidar sensor with a Velodyne VLS-128 reference lidar. Both lidars are mounted on a movable platform allowing to scan a scene from different positions. After PolLidar acquisition is completed, we move the reference lidar through the scene to obtain multiple reference point clouds from different positions. We accumulate the reference point clouds after previous registration with the Iterative-Closest-Point (ICP) algorithm presented in [10] to obtain a dense accumulated lidar map.

**Distance** The lidar map is then used to generate ground truth distances $\mathbf{d}_{\text{gt}}$. To this end, we first transform the lidar map to PolLidar coordinates using the transformation $\mathbf{T} \in \mathbb{R}^{4x4}$. We obtain $\mathbf{T}$ by iteratively aligning reference and PolLidar point clouds for different scenes by means of the ICP algorithm as implemented by [12]. Next, we can extract the ground truth distance by calculating azimuth and elevation angle for each point of the lidar map and then applying bilinear interpolation with the view encoding $\mathbf{V}$ of the PolLidar. As we move the prototype through the scene, the lidar map might have several valid distances per viewing direction that are occluded from one viewpoint but visible from another. This would severely distort the ground truth when applying interpolation to the full lidar map. Thus, before interpolation, we remove hidden points that are invisible from the PolLidar viewpoint by means of [9]. Eventually, we output a ground truth distance map $\mathbf{d}_{\text{gt}} \in \mathbb{R}^{H \times W}$ encoding the ground truth distance for every viewing direction.

**Normals** For generating ground truth normals, we first mesh the lidar map using [8] and then query the mesh at the ground truth point locations for their normals. Compared to ray-tracing with the viewing direction, querying is beneficial as the meshing method often introduces artifacts around object edges enlarging the object. Ray-tracing would produce erroneous normals as the true object is possibly occluded by an enlarged edge. Finally, we output a ground truth normal map $\mathbf{n}_{\text{gt}} \in \mathbb{R}^{H \times W \times 3}$ encoding the ground truth normal for every viewing direction.

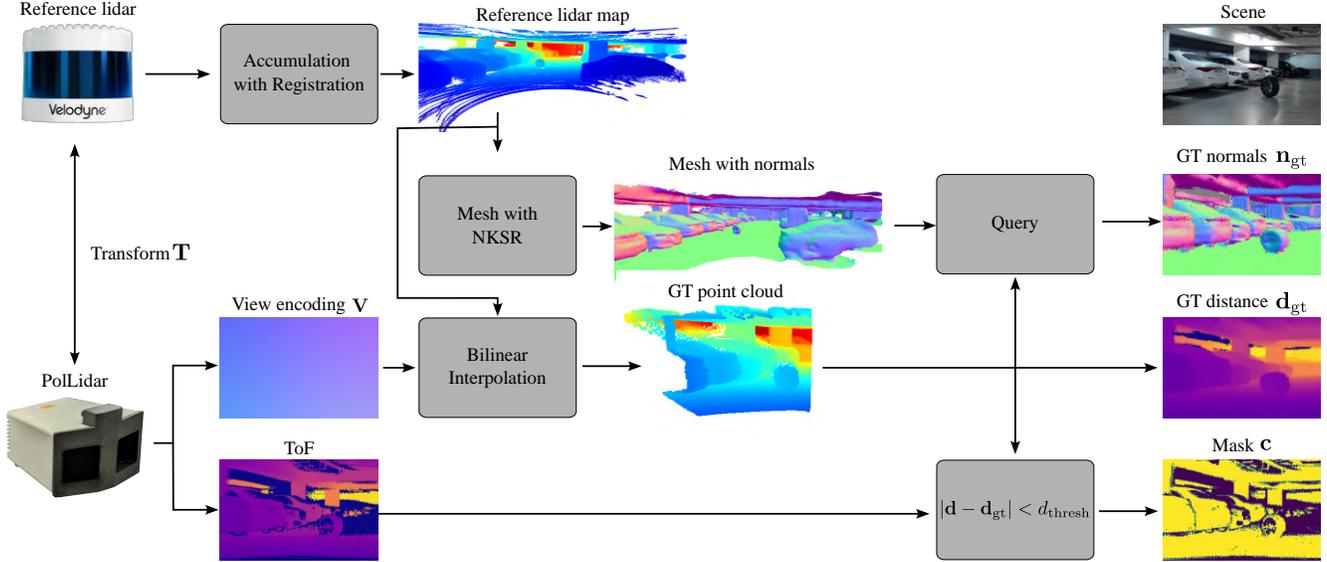

Figure 6. Acquisition of Ground Truth Data. We move PolLidar and reference lidar through a scene to capture dense reference lidar maps. We interpolate the lidar maps with the viewing direction of the PolLidar to extract GT distance $\mathbf{d}_{gt}$ information. For GT normals $\mathbf{n}_{gt}$, we first mesh the lidar map and extract normals from the mesh by querying with the GT point cloud.

**Mask** For certain pixels/view directions, no information can be extracted from the raw wavefront, when e.g. no object was hit, or the measured intensity falls below the noisefloor. We use the ToF map $\mathbf{d}$ extracted from the sensor with conventional peak-finding to exclude these points from training. To this end, we compare if the ToF map and ground truth distance are within a certain bound and define the mask of valid pixels / view directions as

$$\mathbf{c} = \begin{cases} 1 & \text{if } |\mathbf{d} - \mathbf{d}_{GT}| < d_{\text{thresh}} \\ 0 & \text{otherwise} \end{cases}, \tag{16}$$

where $\mathbf{c} \in \mathbb{R}^{H \times W}$ denotes the binary mask of valid view directions and $d_{\text{thresh}}$ is a threshold we define as 0.8m. Furthermore, the mask $\mathbf{c}$ is helpful for eliminating erroneous ground truth. For instance, erroneous ground truth distance are likely to appear around object edges due to accumulation errors and resolution limits of the reference lidar, resulting in a widening of object edges. The mask excludes these points as shown in Fig. 6 for the silhouette of the car.

### 3.2. Acquisition

In the following, we provide further details on how we acquire frames with the PolLidar and describe the settings used. When acquiring a frame, we perform rotating ellipsometry as described in e.g. [4]. To this end, we measure 36 different rotation angle combinations $\theta_i$, where the subscript $i \in \{0, 1, ..., 35\}$ is used to distinguish the different combinations. The resulting stokes vector of the emitted light is visualized in the Poincaré sphere in Fig. 7. As shown, the polarization state of the emitted light is uniformly distributed along the sphere. For the emitter, the HWP is set to $\theta_i^1 = 0$ and the QWP to $\theta_i^2 = 5° \cdot i$. For the receiver, the QWP is rotated to $\theta_i^3 = 25° \cdot i$ and the LP to $\theta_i^4 = 0$.

We opt to capture 10 frames for aggregation per rotation angle $\theta_i$, as we find that it is a good compromise between acquisition time, amount of generated data, and denoising.

Furthermore, we choose a laser power of 600 mV and 900 mV for indoor and outdoor scenes respectively. Due to the high sensitivity of intensity towards the bias voltage, we capture the same scenario with the bias voltages $\{1980, 2000, 2020\}$ mV.

After acquisition with the PolLidar is complete, we capture a reference lidar map to extract ground truth as discussed in Sec. 3.1. To this end, we move the setup through the scenery until we cover a similar area visible from the PolLidar in the initial position.

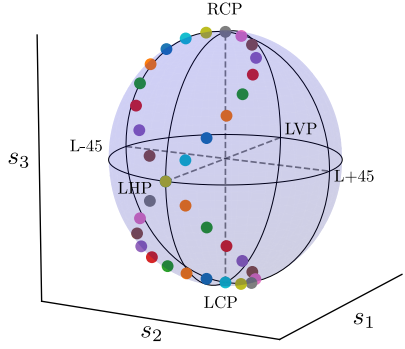

Figure 7. Poincaré sphere visualizing the Stokes vector of the 36 kinds of differently polarized emitted light.

## 4. Reconstruction

The proposed reconstruction approach is a two-step approach. First, we apply classical ellipsometric reconstruction to disentangle the scene from the polarizing optics of the emitter and receiver. From this, we obtain the Mueller matrix $\mathbf{H}$ of the scene. Additional details on this, are provided in Sec. 4.1. Next, we feed this as an input to a neural network that predicts distance offsets and normals. Additional details on the network are provided in Sec. 4.3.

### 4.1. Ellipsometric Reconstruction

Ellipsometric reconstruction is used to disentangle the Mueller matrix of the scene from the Mueller matrices of emitter $\mathbf{P}_i$ and receiver $\mathbf{A}_i$. As discussed in the paper, the additional ellipsometric reconstruction helps the network to learn a better scene reconstruction. In this section, we provide some additional ellipsometric reconstruction results.

In order to recover the Mueller matrix of the scene $\mathbf{H}_{\text{meas}}$, we solve a least-squares optimization problem as defined by

$$\underset{\mathbf{H}_{\text{meas}}}{\text{minimize}} \sum_{i=1}^{N} \left(I_i - [\mathbf{A}_i, \mathbf{H}_{\text{meas}}, \mathbf{P}_i \mathbf{s}_{\text{laser}}]_0\right)^2, \qquad (17)$$

see [2]. We visualize the reconstruction approach in Fig. 8, where the individual elements of the Mueller matrix $\mathbf{H}_{\text{meas}}$ are shown on the right. In order to validate the correctness of the reconstructed Mueller matrix, we then render the intensity image using the lidar forward model and compare it with the measured intensities for the 36 different rotation angles $\theta_i$. This is visualized for the left of Fig. 4.1. On the top, we show the measured intensity for $\theta_0$. On the bottom, we show for two selected pixels, the re-rendered / reconstructed intensities over the 36 different measurements using the polarimetric lidar forward model. We find that the reconstructed intensities are in good agreement with the measured ones. The small deviations are likely due to the inherently noisy measurements. Aside from the disentanglement from the polarization optics, we thus believe that the ellipsometric reconstruction provides additional benefit as an additional denoising step.

### 4.2. Network Details

We present the details of our network architecture in Tab. 3. Specifically, we first use two convolution layers to process the input features. We then use 4 encoder layers to encode the features, each layer consists of a max-pooling and two convolution layers. At the bottleneck, we use 8 transformer layers [5]. At last, we use 4 decoder layers with skip-connection to the prior layer. Finally, we use a 1×1 convolution layer to get a four-channel output, which is the normals and distance, respectively.

### 4.3. Training Details

We evaluate our method on both the simulation data and experimental data. Our simulated Carla dataset consists of 62 different scenes with different IDs. The contents of each scene are generally different. We select 44 scenes and 18 scenes for training and evaluation respectively, leading to 1430 and 539 test frames. We apply different laser biases during training. Specifically, we random sample the bias of our simulated PolLidar from 10 to 900. The intensities and distances for different biases are shown in Fig. 9.

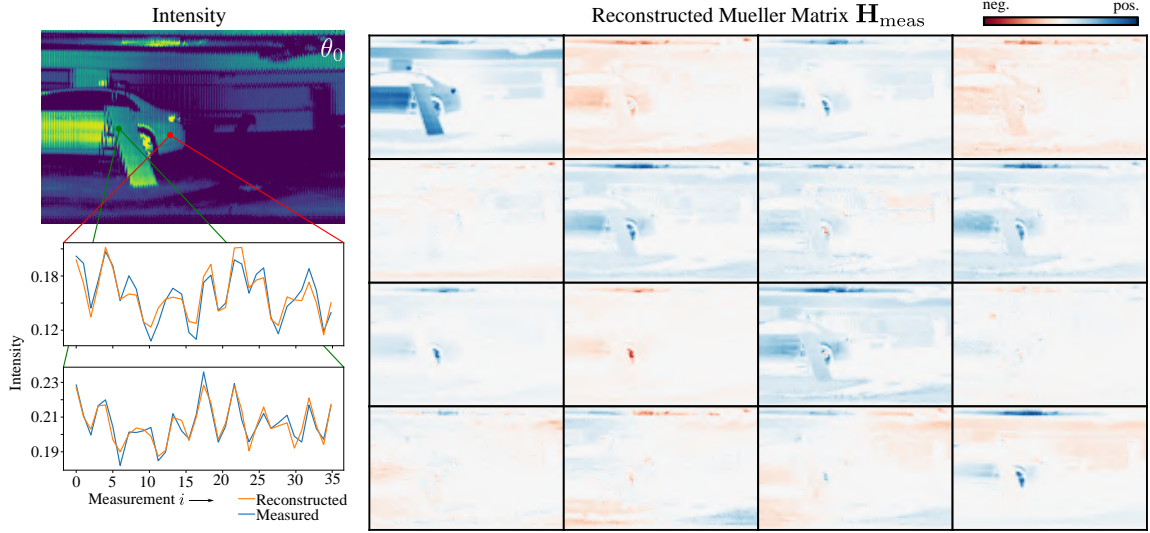

Figure 8. Ellipsometric Reconstruction. We show the individual elements of the reconstructed Mueller matrix on the right. To validate the correctness of the reconstruction, we render the intensities using the discussed polarimetric lidar forward model, as shown on the left. We find agreement between re-rendered / reconstructed and measured intensities. Note that different color scales are applied to each element for better visualization.

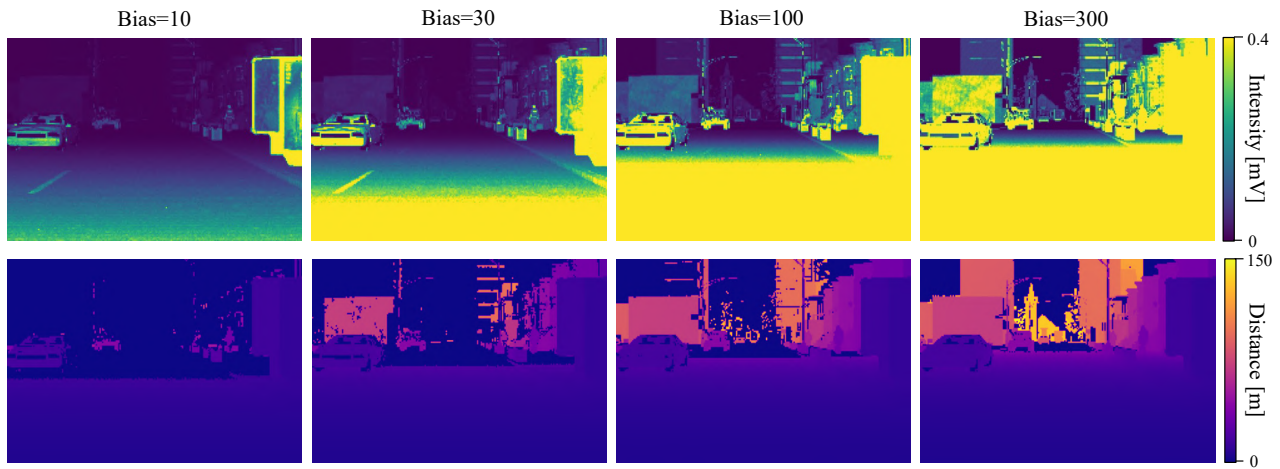

Figure 9. Frames with different biases. When the bias is low, a limited number of points are detected as the intensity falls below the noise floor. When the bias is large, we see saturation effects in regions close to the sensor or in regions of high reflectivity. We use different biases at training time to increase the robustness against saturation and low-intensity readings.

## 5. Additional Quantitative and Qualitative Results

In the following, we provide additional quantitative and qualitative results on real and synthetic data to validate the proposed method further.

### 5.1. Additional Synthetic Results

In Fig. 11, we provide additional synthetic results. Consistent with the main paper, we find that existing SfP methods underperform in low DoP regions. To further validate this, we show the DoP defined as

$$\text{DoP} = \frac{\sqrt{\mathbf{H}_{0,1}^2 + \mathbf{H}_{0,2}^2}}{\mathbf{H}_{0,0}}, \tag{18}$$

| Name | Layer setting | Output dimension |
|---|---|---|
| input conv | Conv $[3 \times 3, 64]$, Instance Normalization, ReLU <br> Conv $[3 \times 3, 64]$, Instance Normalization, ReLU | $H \times W \times 64$ |
| encoder1 | MaxPooling, stride 2 <br> Conv $[3 \times 3, 128]$, Instance Normalization, ReLU <br> Conv $[3 \times 3, 128]$, Instance Normalization, ReLU | $\frac{1}{2}H \times \frac{1}{2}W \times 128$ |
| encoder2 | MaxPooling, stride 2 <br> Conv $[3 \times 3, 256]$, Instance Normalization, ReLU <br> Conv $[3 \times 3, 256]$, Instance Normalization, ReLU | $\frac{1}{4}H \times \frac{1}{4}W \times 256$ |
| encoder3 | MaxPooling, stride 2 <br> Conv $[3 \times 3, 512]$, Instance Normalization, ReLU <br> Conv $[3 \times 3, 512]$, Instance Normalization, ReLU | $\frac{1}{8}H \times \frac{1}{8}W \times 512$ |
| encoder4 | MaxPooling, stride 2 <br> Conv $[3 \times 3, 512]$, Instance Normalization, ReLU <br> Conv $[3 \times 3, 512]$, Instance Normalization, ReLU | $\frac{1}{16}H \times \frac{1}{16}W \times 512$ |
| attention | Transformer Block $\times 8$ | $\frac{1}{16}H \times \frac{1}{16}W \times 512$ |
| decoder4 | Concat [encoder4 outputs, attention outputs] <br> Bilinear Upsampling with scale 2 <br> Conv $[3 \times 3, 512]$, Instance Normalization, ReLU <br> Conv $[3 \times 3, 512]$, Instance Normalization, ReLU | $\frac{1}{8}H \times \frac{1}{8}W \times 512$ |
| decoder3 | Concat [encoder3 outputs, decoder4 outputs] <br> Bilinear Upsampling with scale 2 <br> Conv $[3 \times 3, 256]$, Instance Normalization, ReLU <br> Conv $[3 \times 3, 256]$, Instance Normalization, ReLU | $\frac{1}{4}H \times \frac{1}{4}W \times 256$ |
| decoder2 | Concat [encoder2 outputs, decoder3 outputs] <br> Bilinear Upsampling with scale 2 <br> Conv $[3 \times 3, 128]$, Instance Normalization, ReLU <br> Conv $[3 \times 3, 128]$, Instance Normalization, ReLU | $\frac{1}{2}H \times \frac{1}{2}W \times 128$ |
| decoder1 | Concat [encoder1 outputs, decoder2 outputs] <br> Bilinear Upsampling with scale 2 <br> Conv $[3 \times 3, 64]$, Instance Normalization, ReLU <br> Conv $[3 \times 3, 64]$, Instance Normalization, ReLU | $H \times W \times 64$ |
| output conv | Conv $[1 \times 1, 4]$ | $H \times W \times 4$ |

Table 3. Details of our network architecture. "Concat" operation means that we concatenate two elements along the channel dimension.

where the subscript denotes the respective index of the Mueller matrix **H**. We show the DoP for the two selected scenes of Fig. 5 in the main paper in Fig. 10. When normals and viewing direction are aligned, as visualized in the two right columns, the DoP is low. This is true for e.g. buildings and parts of the car that face the sensor. Contrarily, the side or hood of the car is for instance a high DoP region as viewing direction and normal are almost perpendicular. Consequently, Baek et al. [2] are unable to reconstruct normals in these low DoP regions, as shown in Fig. 11. For high DoP regions, however, satisfying performance is achieved.

We also compare the normal reconstruction to PCA as a point-cloud based method that considers a neighborhood of points. This method performs well in areas with flat geometry and high point density but degrades significantly at long ranges, e.g., cars in far distances, and geometry transition regions, e.g., the area between road and car. The proposed method leverages the additional cues from polarization to resolve normals in regions with sparse points and in transition regions. Furthermore, the proposed method achieves satisfying reconstruction results for regions with little polarization information by taking a local neighborhood into account.

## 5.2. Additional Real-World Results

We provide additional qualitative results from the real-world dataset in Fig. 12. We find similar trends as in the main paper. The proposed method consistently outperforms PCA in areas where the point cloud is sparse. This is visible in the zoom-ins on the fine structures, e.g., car roofs, where the neighborhood is too sparse for PCA to reconstruct correct surface normals. This is visible for the roof structure in the fourth row or the car on the left in the fifth row. The point cloud visualization also

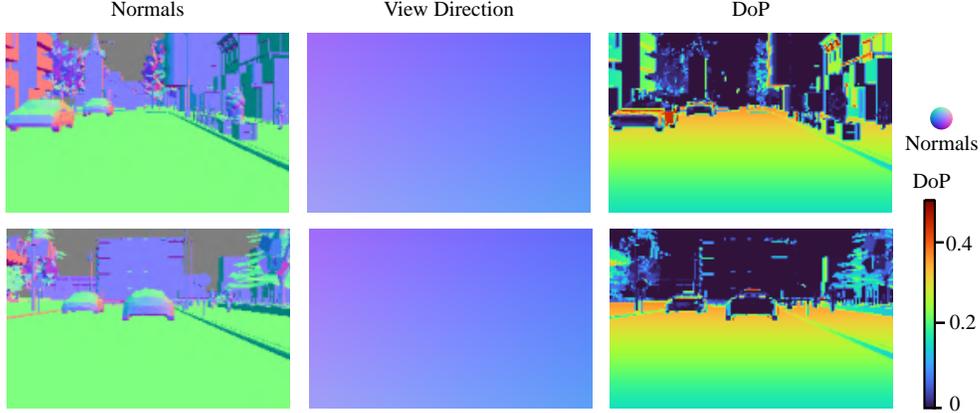

Figure 10. Degree-of-Polarization (DoP): On the right, we show the DoP for the two scenes of Fig. 5 in the main paper. When comparing with normals and viewing direction, we find low DoP regions on objects where normals and viewing direction align. This is problematic for other SfP reconstruction approaches as little polarization cues are present in these areas for a successful reconstruction.

reveals the benefit of the distance reconstruction compared to conventional argmax-methods in the context of reconstructing fine details. This is pronounced in the fourth row of Fig. 12 for the bumper/headlight area of the car on the left. The proposed method is able to reconstruct the pocket in the chassis where the headlight is located correctly. This detail is lost in the PCA results, as here, conventional peak-finding is applied. Furthermore, our method provides high-quality reconstruction results in the area around the grill of the car. Additionally, we show data in night-time conditions where the PolLidar expectedly shows similar reconstruction results as during day-time conditions.

### 5.3. Additional Distance-Binned Evaluation of Normal Reconstruction

To further analyze the advantage of our approach over the existing baseline in regions of low point density, we analyze the mean angular error for normal reconstruction and bin the results per distance bin. Specifically, this analysis allows us to study further distances in more detail, where point distance due to constant angular sampling increases further, making PCA fail due to the missing neighborhood. Quantitative results are visualized in Fig. 13, showing in (a) an increased performance due to polarization in close distance by approx. factor of two compared to PCA. Thereby, for further distances PCA degenerates by 75% being outperformed by a factor of 2.5. Furthermore, our method is able to cope with sparser point clouds and shows a substantially lower dependency of reconstruction performance on distance. In Fig. 13(b), we plot the relative mean angular error between PCA and the proposed method. We see that the gain in reconstruction quality is closely related to the distance.

### 5.4. Additional Metrics for Distance Evaluation

To evaluate the distance estimation, we compare against the conventional argmax-peak-finding typically performed directly on the device by low-level electronics [3]. We list all evaluated metrics on the distance reconstruction in Tab. 4 and Tab. 5. In addition to the results in the main paper, we present here median and RMSE distance errors. We find that our approach outperforms conventional peak-finding on these metrics by large margins on both synthetic and real data, outperforming previous results by 41% on synthetic data and 17% on real-world data for mean absolute distance error. The comparatively lower gain in real-world data is likely related to the quality of the ground truth. For synthetic data, we have perfect ground truth as we have control over the entire rendering pipeline including the underlying geometry. However, for the real-world data, we are limited by our applied geometry reconstruction described in Sec. 3.1. This has inherent sensor noise obfuscating distance and inaccuracies of the accumulation of our ground-truth though pose estimation errors, beam divergence broadening object dimensions, and many more. Consequently, the median error is increased for both conventional and proposed distance estimation methods. In contradiction to that is the higher RMSE in the synthetic data which can be explained with outliers in the distance error. More specifically, if the conventional peak-finding method selects the wrong peak or misses the peak altogether, e.g., due to low intensities close to the noise floor, the distance error will be of many meters for both the conventional and proposed method, thus increasing the RMSE significantly. For the real data, the effect of increased RMSE is not as prominent, as we apply the mask dependent on $d_{\text{thresh}}$ effectively suppressing these outliers.

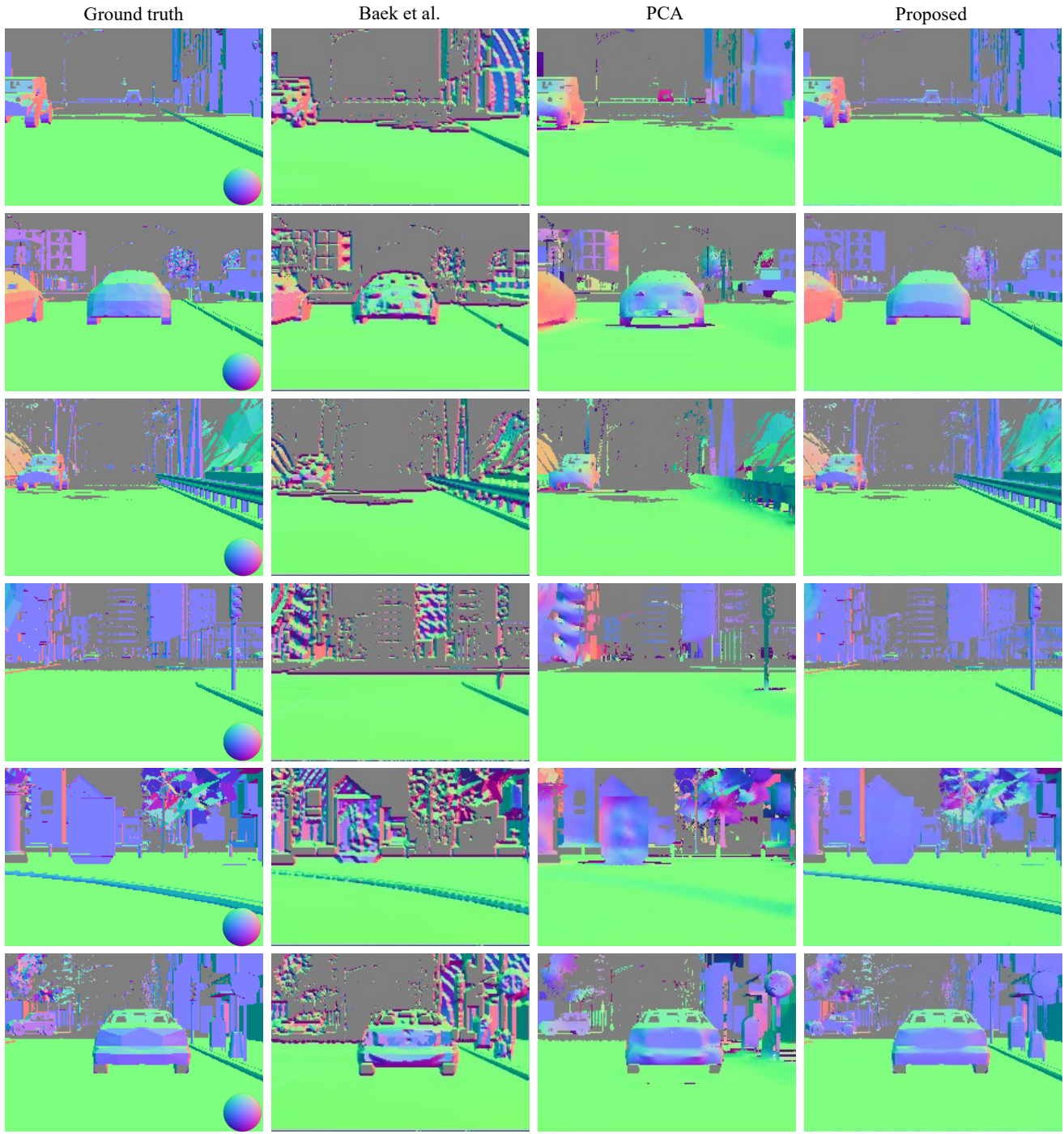

Figure 11. Additional Synthetic Results. Baek et al. [2] is unable to reconstruct normals in areas with low DoP, e.g., walls of buildings facing the sensor. PCA [13] applied in this setting is strongly dependent on point cloud density. This is visible for e.g. the poles in far distances. The proposed approach leverages polarization cues to reconstruct normals in sparse regions and is robust against low DoP areas.

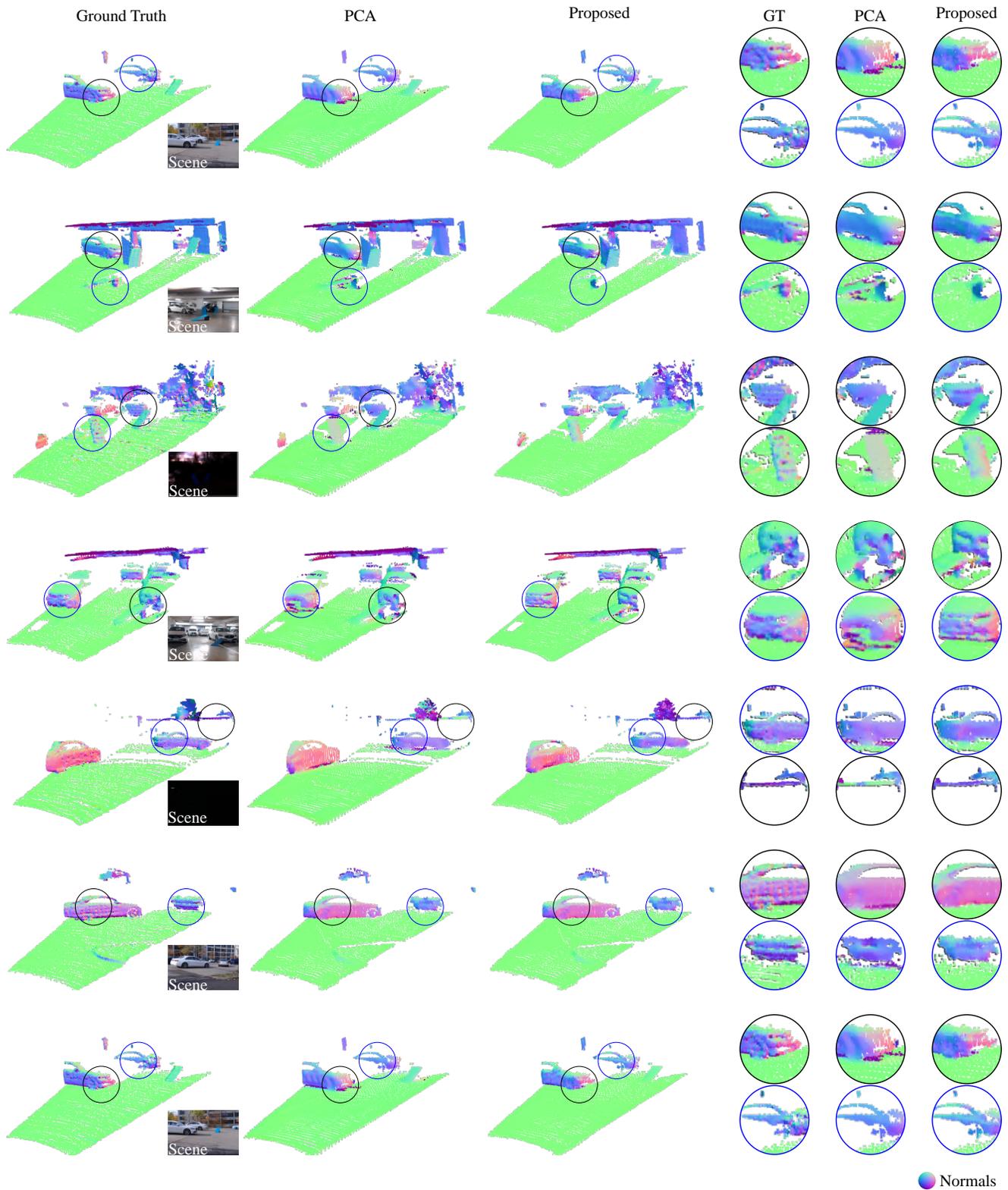

Figure 12. Additional Real-world Results. We show day- and night-time automotive scenes. PCA generates high-quality surfaces in areas with high point density. However, for fine details or objects far away, where the point cloud is sparse, the reconstruction is of low quality. The proposed method, however, leverages the polarization cues in these areas and thus outperforms PCA in regions of low-point density. This can be seen in the zoom-ins shown on the right, e.g. bumper area of the car in the fourth row or car roofs in the fifth and sixth rows.

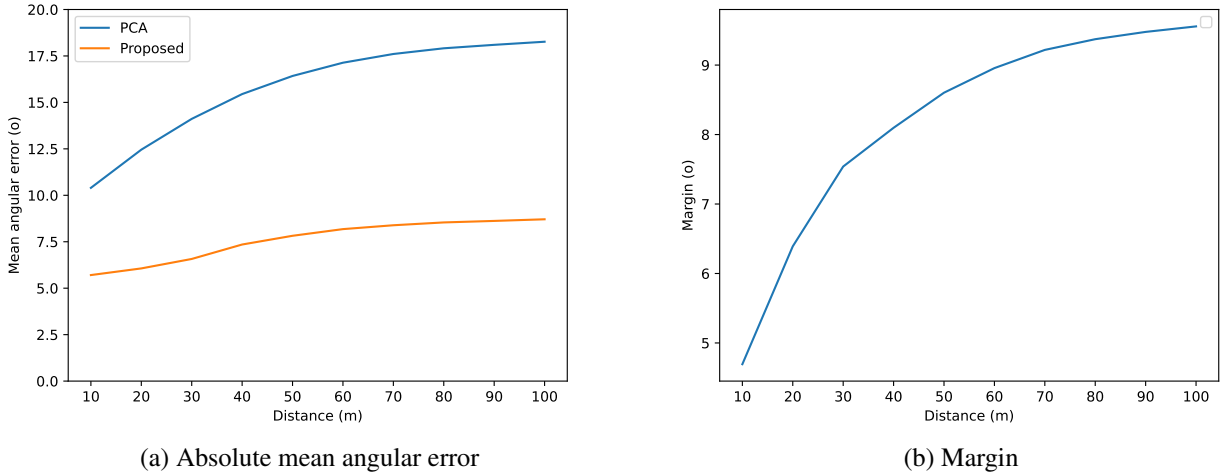

(a) Absolute mean angular error

(b) Margin

Figure 13. Mean angular error dependent on distance. In (a), PCA achieves reasonable performance for close regions as the point clouds are more dense here. However, as the distance increases, the mean angular error of PCA increases rapidly. In (b), we plot of the margin between PCA and the proposed approach. We can see the performance gain is closely correlated with the distance.

| Method | Distance Error [m] ↓ | | |
|---|---|---|---|
| | Mean | Median | RMSE |
| Conventional | 0.32 | 0.12 | 3.25 |
| Proposed | **0.19** | **0.03** | **3.06** |

Table 4. Quantitative Distance Error on Synthetic Data. Our approach outperforms existing baselines on all evaluated distance metrics. Conventional distance estimation is limited by the temporal resolution of the sensor. Our approach leverages the wavefront information, thus outperforming the conventional distance estimation approach.

| Method | Distance Error [m] ↓ | | |
|---|---|---|---|
| | Mean | Median | RMSE |
| Conventional | 0.24 | 0.20 | 0.29 |
| Proposed | **0.20** | **0.18** | **0.26** |

Table 5. Quantitative Distance Error on Real-world Data. Due to noisier ground truth and sensor imperfections, the overall error is slightly larger compared to synthetic data, while confirming the trend from the synthetic evaluation. The proposed method relies on wavefront data and point neighborhoods, outperforming the conventional approach.

# References

[1] Gary A. Atkinson and Edwin R. Hancock. Recovery of surface orientation from diffuse polarization. *IEEE Trans. Image Process.*, 15(6):1653–1664, 2006. 4, 5

[2] Seung-Hwan Baek and Felix Heide. All-photon polarimetric time-of-flight imaging. *Proceedings of the IEEE Conference on Computer Vision and Pattern Recognition (CVPR)*, 2022. 5, 10, 12, 14

[3] Behnam Behroozpour, Phillip AM Sandborn, Ming C Wu, and Bernhard E Boser. Lidar system architectures and circuits. *IEEE Communications Magazine*, 55(10):135–142, 2017. 13

[4] Edward Collett. Field guide to polarization. Spie Bellingham, WA, 2005. 6, 9

[5] Alexey Dosovitskiy, Lucas Beyer, Alexander Kolesnikov, Dirk Weissenborn, Xiaohua Zhai, Thomas Unterthiner, Mostafa Dehghani, Matthias Minderer, Georg Heigold, Sylvain Gelly, et al. An image is worth 16x16 words: Transformers for image recognition at scale. *arXiv preprint arXiv:2010.11929*, 2020. 10

[6] Felix Goudreault, Dominik Scheuble, Mario Bijelic, Nicolas Robidoux, and Felix Heide. Lidar-in-the-loop hyperparameter optimization. 2023. 2, 7

[7] Eric Heitz. Understanding the masking-shadowing function in microfacet-based brdfs. *Journal of Computer Graphics Techniques*, 3(2):32–91, 2014. 7

[8] Jiahui Huang, Zan Gojcic, Matan Atzmon, Or Litany, Sanja Fidler, and Francis Williams. Neural kernel surface reconstruction. In *Proceedings of the IEEE/CVF Conference on Computer Vision and Pattern Recognition*, pages 4369–4379, 2023. 8

[9] Sagi Katz, Ayellet Tal, and Ronen Basri. Direct visibility of point sets. In *ACM SIGGRAPH 2007 papers*, pages 24–es. 2007. 8

[10] Ignacio Vizzo, Tiziano Guadagnino, Benedikt Mersch, Louis Wiesmann, Jens Behley, and Cyrill Stachniss. Kiss-icp: In defense of point-to-point icp–simple, accurate, and robust registration if done the right way. *IEEE Robotics and Automation Letters*, 8(2):1029–1036, 2023. 8

[11] Bruce Walter, Stephen R Marschner, Hongsong Li, and Kenneth E Torrance. Microfacet models for refraction through rough surfaces. In *Proceedings of the 18th Eurographics conference on Rendering Techniques*, pages 195–206, 2007. 7

[12] Qian-Yi Zhou, Jaesik Park, and Vladlen Koltun. Open3D: A modern library for 3D data processing. *CoRR*, abs/1801.09847, 2018. 8

[13] Yufan Zhu, Weisheng Dong, Leida Li, Jinjian Wu, Xin Li, and Guangming Shi. Robust depth completion with uncertainty-driven loss functions. In *Proceedings of the AAAI Conference on Artificial Intelligence*, volume 36, pages 3626–3634, 2022. 14



\end{document}